\documentclass[10pt,twocolumn,letterpaper]{article}

\usepackage[pagenumbers]{wacv} %

\usepackage{graphicx}
\usepackage{amsmath}
\usepackage{amssymb} %
\usepackage{booktabs}

\usepackage{mathtools}
\usepackage{placeins}
\usepackage{stmaryrd} %
\usepackage[table]{xcolor} %
\usepackage{verbatim}        %
\usepackage{cprotect}        %
\usepackage{float}           %
\usepackage{cellspace, makecell} %

\usepackage{graphicx} %
\usepackage{tikz} %

\usetikzlibrary{positioning,shapes.misc,fit} %

\usepackage[pagebackref,breaklinks,colorlinks]{hyperref}

\usepackage[capitalize]{cleveref}
\crefname{section}{Sec.}{Secs.}
\Crefname{section}{Section}{Sections}
\Crefname{table}{Table}{Tables}
\crefname{table}{Tab.}{Tabs.}

\definecolor{grey}{rgb}{0.3, 0.3, 0.3}
\definecolor{green}{rgb}{0.22, 0.396, 0.082}
\definecolor{purpleblue}{rgb}{0.51, 0.298, 0.784}
\definecolor{dblue}{rgb}{0.231, 0.345, 0.62}
\definecolor{dyellow}{rgb}{0.89, 0.729, 0.153}

\def\wacvPaperID{1703} %
\def\confName{WACV}
\def\confYear{2024}

\begin{document}

\newcommand{\titleabbreviation}{\mbox{MoRF}}
\title{MoRF: Mobile Realistic Fullbody Avatars from a Monocular Video}

\author{
{Renat Bashirov\textsuperscript{1}}
\and{Alexey Larionov\textsuperscript{1}}
\and{Evgeniya Ustinova\textsuperscript{1}} 
\and{Mikhail Sidorenko\textsuperscript{1}}
\and{David Svitov\textsuperscript{1}} 
\and{Ilya Zakharkin\textsuperscript{2}}\thanks{Contributed during employment at Samsung Research}
\and{Victor Lempitsky\textsuperscript{3}}\footnotemark[1]
\and
\and\textsuperscript{1}Samsung Research
\hspace{0.3cm}\textsuperscript{2}ZERO10
\hspace{0.3cm}\textsuperscript{3}Cinemersive Labs
}
\maketitle

\begin{abstract}
   We present a system to create \textbf{Mo}bile \textbf{R}ealistic \textbf{F}ullbody (\titleabbreviation{}) avatars. \titleabbreviation{} avatars are rendered in real-time on mobile devices, learned from \mbox{monocular} videos, and have high realism. We use \mbox{SMPL-\textbf{X}} as a proxy geometry and render it with DNR (neural texture and image-2-image network). We improve on prior work, by overfitting per-frame warping fields in the neural texture space, allowing to better align the training signal between different frames. We also refine \mbox{SMPL-X} mesh fitting procedure to improve the overall avatar quality. In the comparisons to other monocular video-based avatar systems, \titleabbreviation{} avatars achieve higher image sharpness and temporal consistency. Participants of our user study also preferred avatars generated by \titleabbreviation{}.
\end{abstract}

\section{Introduction}
\label{sec:intro}

This work focuses on fullbody human avatars with fast rendering, realism and the ease of acquisition. For our system we: 1) use short (\ie minute-long) monocular videos as input, 2) train for a few hours on a single GPU, 3) provide sharp texture of the resulting avatar, and 4) can render in real-time on mobile devices. As far as we know, despite impressive progress, existing systems fall short of these criteria. Specifically, some rely on multi-view videos \cite{mixture-volum-primitives:2021, dracon:2022}, or a depth sensor \cite{robust-3d-portraits-in-seconds-rgbd:2020, metaavatar-rgbd:2021}, some leverage implicit representations whose rendering process is far from real-time especially on mobile devices \cite{phorhum,neural-body:2021, nerf:2020, animnerf:2021, neural-actor, eva3d:2022}, whilst others take days to converge on multiple GPUs \cite{dance-in-the-wild:2021,dynamic-humans:2022}.

{%
\setlength{\belowcaptionskip}{-6pt}
\begin{figure}
\centering
    \setlength{\tabcolsep}{0pt} %
    \setlength{\arrayrulewidth}{0pt}
    \newcommand{\mysmartphonefakeheight}{6cm}
    \begin{tikzpicture}[every node/.style={inner sep=0,outer sep=0}]
        \clip [rounded corners=0.3cm] (0,0) rectangle coordinate (centerpoint) ++(\columnwidth-0.15cm,\mysmartphonefakeheight);
        \node(current bounding box.south west) [anchor=south west] (bb) at (0,0) {\scalebox{1}[1]{\includegraphics[height=\mysmartphonefakeheight,trim={110 40 100 200},clip]{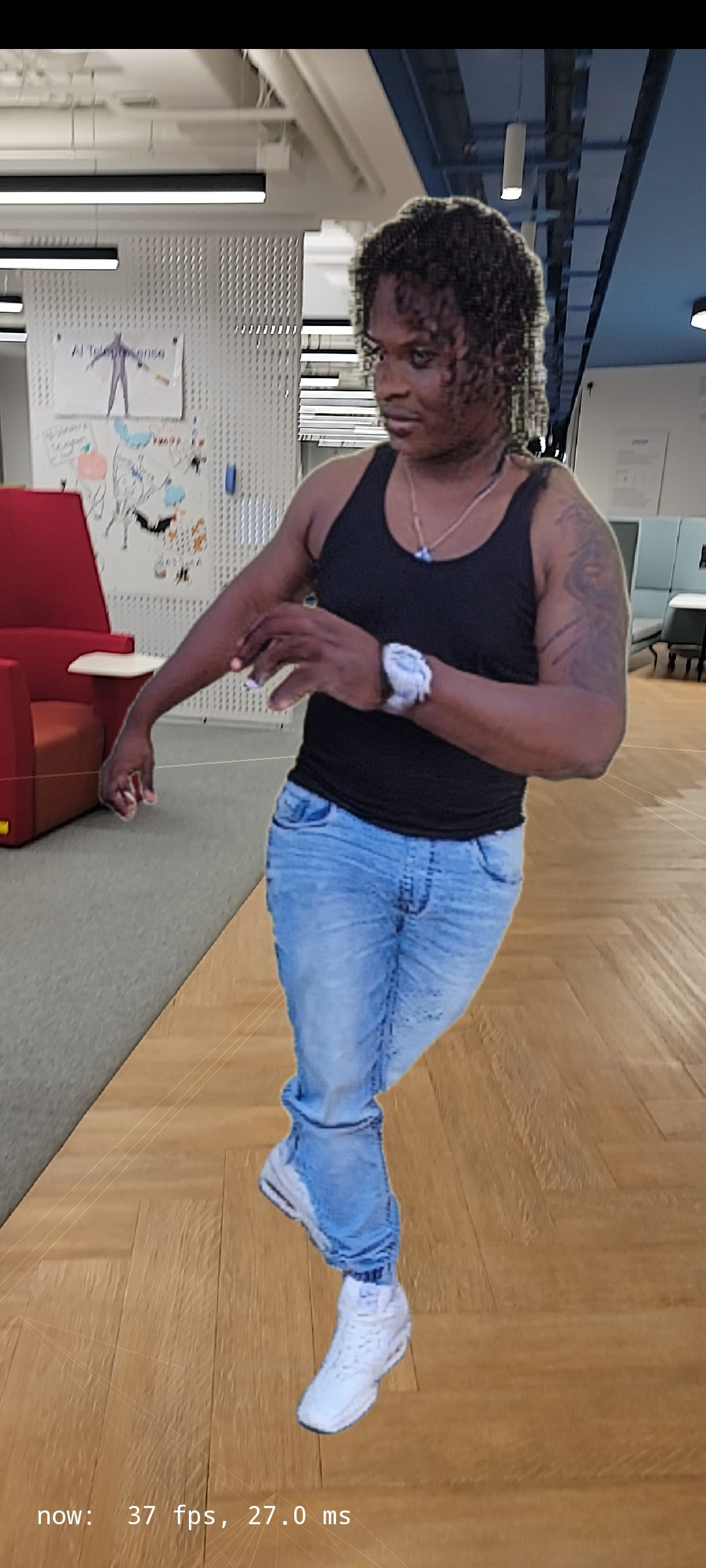}}};
        \node [right=1pt of bb] (david) {\includegraphics[height=\mysmartphonefakeheight,trim={150 40 20 400},clip]{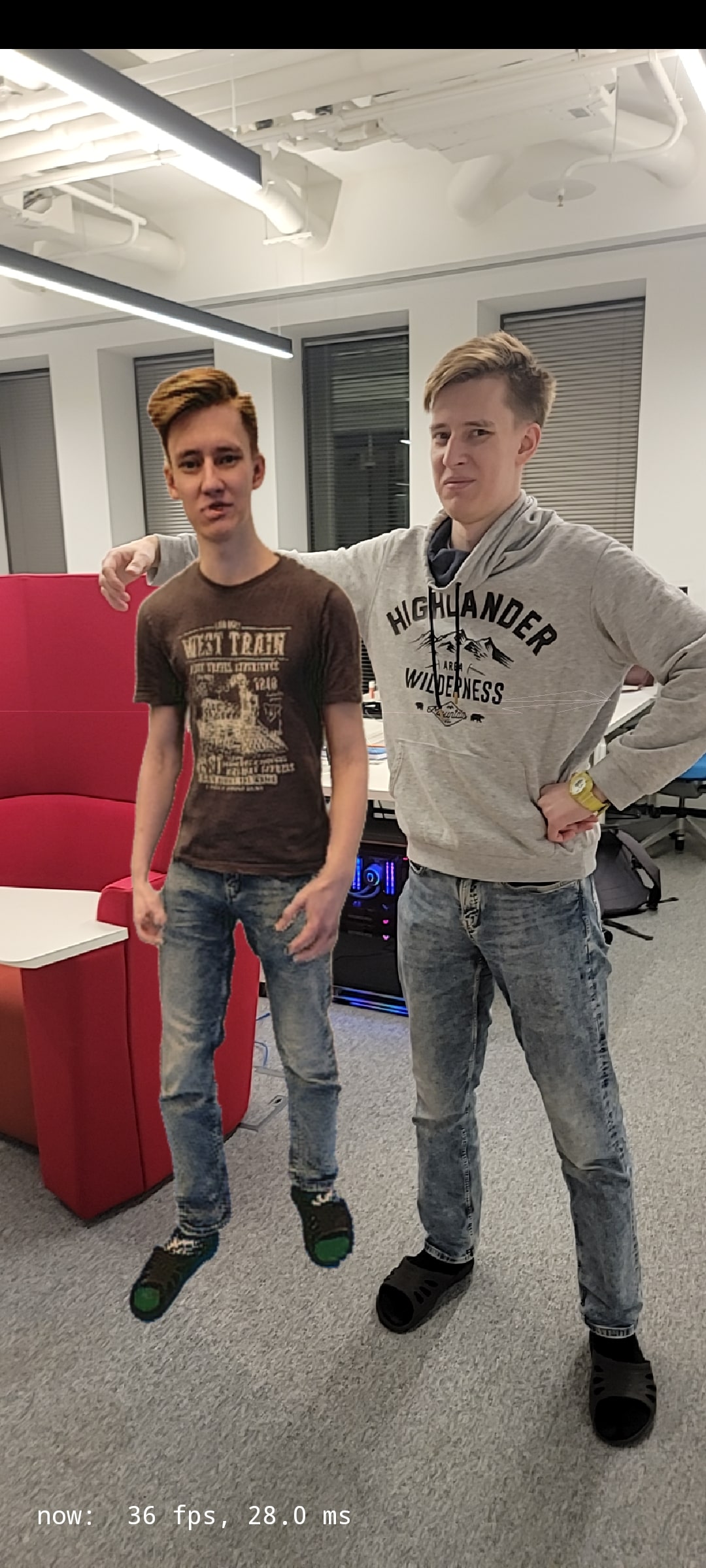}};
        \node [right=1pt of david] (minsoo) {\includegraphics[height=\mysmartphonefakeheight,trim={20 00 0 220},clip]{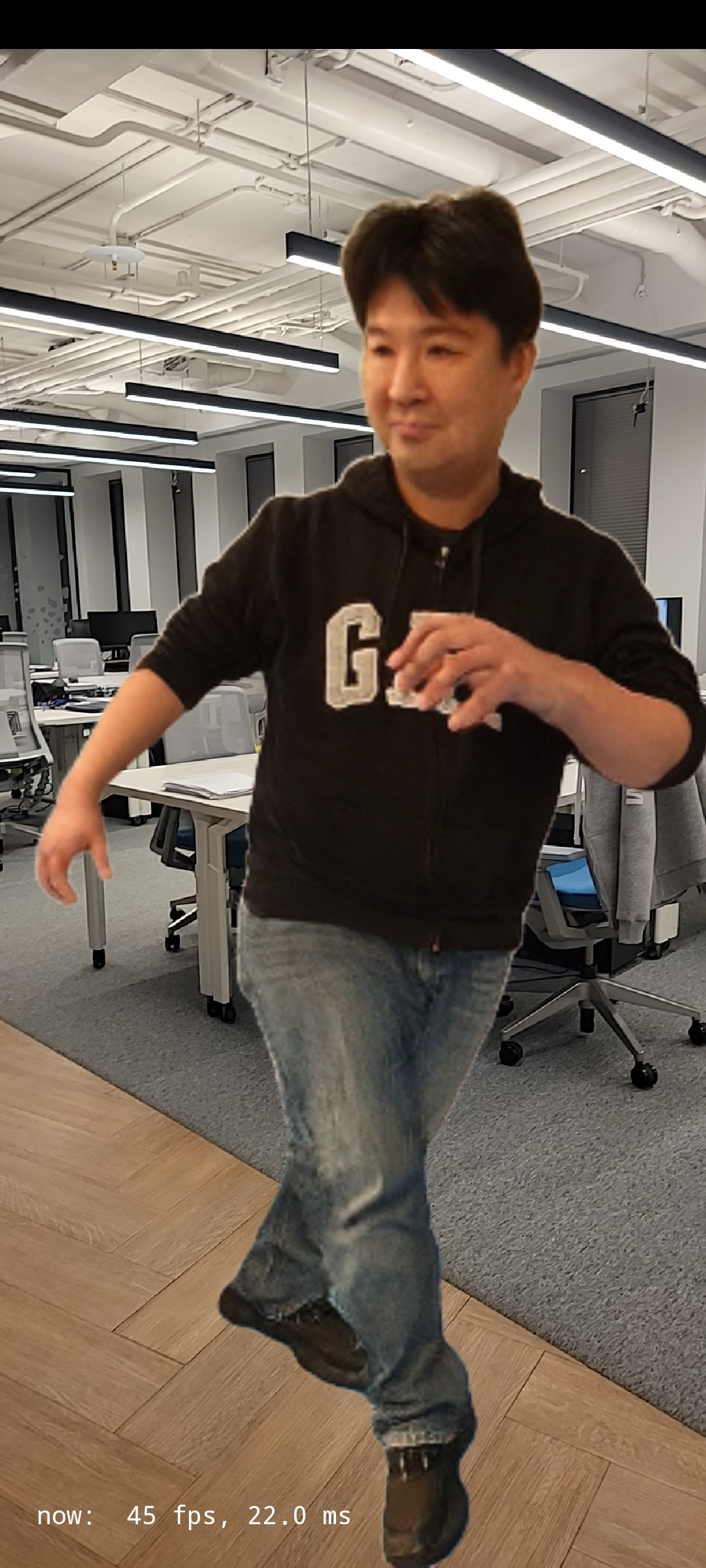}};
    \end{tikzpicture}
\caption{Three MoRF avatars (and a real person in gray) rendered on a mobile phone in realtime (at least 30 frames per second) in the augmented reality mode. Each of the avatars was created from a monocular video, while the poses and the camera positions of these examples differ from the training frames.} %
\label{fig:smartphone}
\end{figure}
}

We therefore present a system for \textbf{Mo}bile \textbf{R}ealistic \textbf{F}ullbody (\titleabbreviation{}) avatars. Our avatars are realistic and exhibit temporal consistency of generated images, and they can be rendered in realtime at 30 FPS on mobile devices (Fig.~\ref{fig:smartphone}) with Qualcomm Snapdragon 888 SoC (2020's flagship SoC for mobile phones). Furthermore, they can be acquired from monocular videos in a few hours on a single NVIDIA RTX 3090 GPU. We rely on the deferred neural rendering (DNR) idea \cite{dnr:2019}, like in numerous prior avatar systems~\cite{anr:2021,dynamic-garments:2021,dynamic-humans:2022,dressing-avatars:2022}, to achieve rendering speed and realism. DNR represents 3D objects using coarse 3D geometry and \textit{neural textures}, which are supplemented by a rendering image-2-image translation network to produce realistic images. For geometric modeling, we use the well-established SMPL-X body model~\cite{SMPL-X:2019}.

{%
\setlength{\belowcaptionskip}{-6pt}
\begin{figure*}[t]
    \centering
    \includegraphics[width=1.0\textwidth]{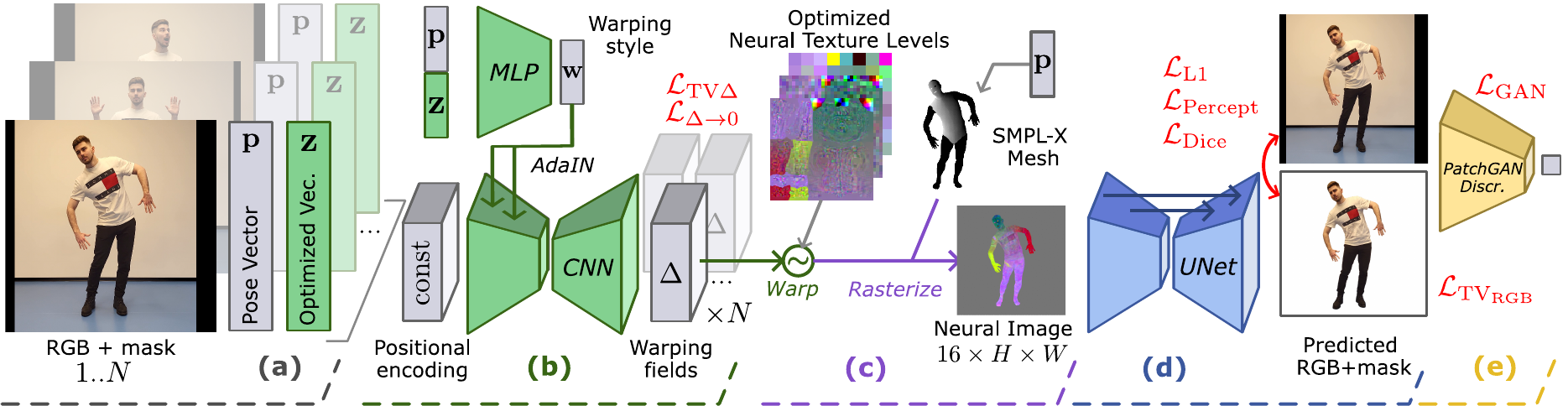}

    \caption{The proposed architecture. %
    The colored blocks highlight end-to-end optimized parts. \textbf{\textcolor{grey}{(a)}} The frames, pose vectors, and segmentation masks (not shown) are all inputs to the modeling process. In addition, for each frame, a latent texture warping vector is introduced. \textbf{\textcolor{green}{(b)}} A texture warping field is computed for each frame using a convolutional architecture and the respective warping vector.  \textbf{\textcolor{purpleblue}{(c)}}  The neural texture associated with the avatar is warped according to the predicted field and is superimposed over a posed SMPL-X mesh. \textbf{\textcolor{dblue}{(d)}} The rendering network predicts an RGB image as well as a segmentation mask. This network also inpaints out-of-mesh details, \eg garments and hair. \textbf{\textcolor{dyellow}{(e)}} We supervise with pixelwise, perceptual, adversarial losses on predicted images; Dice loss on predicted segmentations.}  
    \label{fig:arch}
\end{figure*}
}

We discovered that naively implementing the DNR approach results in underfitting blurriness or overfitting flickering. To address them, we introduce a new component into the DNR learning process that we call the \textit{neural texture warping}. For each training image/frame, it fits a warping field for the neural texture, thus compensating between-frame mesh fit misalignment and facilitating the task for image-2-image UNet renderer. 

The warping fields are only used for training to compensate for mesh fitting imperfections, this produces sharp and static texture for the resulting avatar. \textit{We don't solve the task of clothing modelling/movement} and the warping fields are discarded after the training process has converged. %

We evaluate our approach on 3 datasets: self-captured, ZJU-MoCap \cite{neural-body:2021,fang2021mirrored} and People Snapshot \cite{people-snapshot}. Self-captured and ZJU-MoCap provide ground truth for novel body and camera poses respectively. We also conduct user studies for avatar rendering quality and for mesh fitting refinements ablation. We compare against InstantAvatar~\cite{jiang2023instantavatar}, HF-Avatar~\cite{hf-avatar}, \mbox{Anim-NeRF}~\cite{animnerf:2021}, \mbox{HumanNeRF}~\cite{humannerf:2022}, \mbox{StylePeople}~\cite{stylepeople}, \mbox{ANR}~\cite{anr:2021} and \mbox{NeuMan}~\cite{neuman}.

\section{Related Work}\label{sec:related}

Modeling geometry of dynamic non-rigid scenes is considered in a number of recent approaches either for capturing human actors \cite{driving-signal-aware, neuralgif:2021, avatarcap:2022, tava:2022,neural-actor, power-of-points-humans-clothing:2021, habermann2021real, arch, arch++, humannerf:2022} or general scenes \cite{neural-volumes:2019, pumarola2020d, nerfies:2021}. Towards this end,  3D scans are required for supervision in \cite{arch, arch++, power-of-points-humans-clothing:2021, neuralgif:2021, avatarcap:2022}  to learn rigged human geometry. Likewise, \cite{neural-volumes:2019, neural-actor,prokudin2021smplpix, habermann2021real, driving-signal-aware,  drivable-volumetric-avatars:2022, tava:2022} utilize multi-view data to capture appearance and produce photo-realistic avatars under arbitrary viewpoints and in arbitrary poses. Several methods \cite{humannerf:2022, pumarola2020d, nerfies:2021} use monocular videos but allow free-viewpoint rendering only. 

 Some of the monocular fullbody avatar methods model the human geometry implicitly \cite{neural-actor, animnerf:2021, humannerf:2022}, others  output the classical mesh+texture format \cite{video-avatar, octopus, alldieck2018detailed, hf-avatar, stylepeople, anr:2021, svitov2023dinar}. Similarly to StylePeople~\cite{stylepeople} and ANR~\cite{anr:2021}, our method is based on DNR~\cite{dnr:2019} allowing for efficient convolutional rendering on a mobile device. However, StylePeople leaves modeling of geometry imperfections and frame-to-frame misalignments to the neural renderer, while ANR handles these issues by limiting the number of frames used for the neural texture optimization. HF-Avatar~\cite{hf-avatar} uses the synthetic images as supervision for the neural texture, because of their consistency and good alignment with the geometry.
\cite{alldieck2018detailed} proposes a texture merging optimization procedure to prevent texture averaging among different views. Our neural texture warping allows for consistent learning of the neural texture  without training on synthetic data \cite{hf-avatar}, excluding most frames from texture learning~\cite{anr:2021}, or over-fitting the neural renderer to unmodeled variations~\cite{stylepeople}. 

 Many of the mentioned works introduce learned sub-modules \cite{neural-volumes:2019, habermann2021real, neural-actor, neuralgif:2021, avatarcap:2022, tava:2022, hf-avatar,humannerf:2022} or pipeline stages \cite{alldieck2018detailed, video-avatar} to estimate additional geometry warping in 3D canonical (unposed or view-agnostic) space. Likewise, \cite{neuralgif:2021, avatarcap:2022, tava:2022, neural-actor, humannerf:2022} parametrize the 3D warping fields or displacements \cite{power-of-points-humans-clothing:2021, hf-avatar} as neural networks conditioned on human pose to facilitate learning of complex pose-dependent geometry. 
 Methods for general scenes leverage other variants of conditioning. For example\ \cite{nerfies:2021, pumarola2020d}, that rely on monocular videos, condition 3D warping fields on the point in time and learnable latent of a training frame respectively, while  \cite{neural-volumes:2019} introduce view-conditional warpings as they use multi-view capture.  
 The way we introduce warping in this work is different to all above-referenced approaches, as our learnable warping operates in the canonical 2D texture space. Operating in the texture space naturally fits the DNR approach that we are based upon. %

\section{Method}\label{sec:method}
Our approach aims to learn a rigged full-body human avatar using a relatively short (e.g.~a minute-long) monocular video. We begin by briefly describing deferred neural rendering framework, as it is at the core of our system. We then introduce the neural texture warping that facilitates the training process. We also mention important architectural decisions that contribute to the avatars quality.

\subsection{Recap: Deferred Neural Rendering}

Deferred neural rendering (DNR) \cite{dnr:2019} models scenes  by combining a geometric proxy defined by a triangular mesh, a neural texture $\bf T$ (with $C$ channels), and an image-to-image rendering network (renderer) $\mathcal{R}$ with a convolutional architecture. Novel views for certain camera parameters are synthesized in two steps. The first is to project the mesh onto the camera view, while superimposing the texture on the mesh using texture mapping. The resulting $C$-channeled image can then be ``translated'' into an RGB image (plus optionally a mask) by the rendering network. During training/fitting to a given dataset of images with known corresponding camera parameters, the latent texture $\bf T$ and the parameters of the rendering network $\mathcal{R}$ are jointly optimized to minimize losses between training images and images predicted by $\mathcal{R}$.

\subsection{Neural Texture Warping}\label{sec:method:warping}

{%
\setlength{\belowcaptionskip}{-6pt}
\begin{figure}[t]
\centering
    \subfloat[\centering Image space \label{fig:misalign:rgb}]
    {\includegraphics[height=6.0cm]{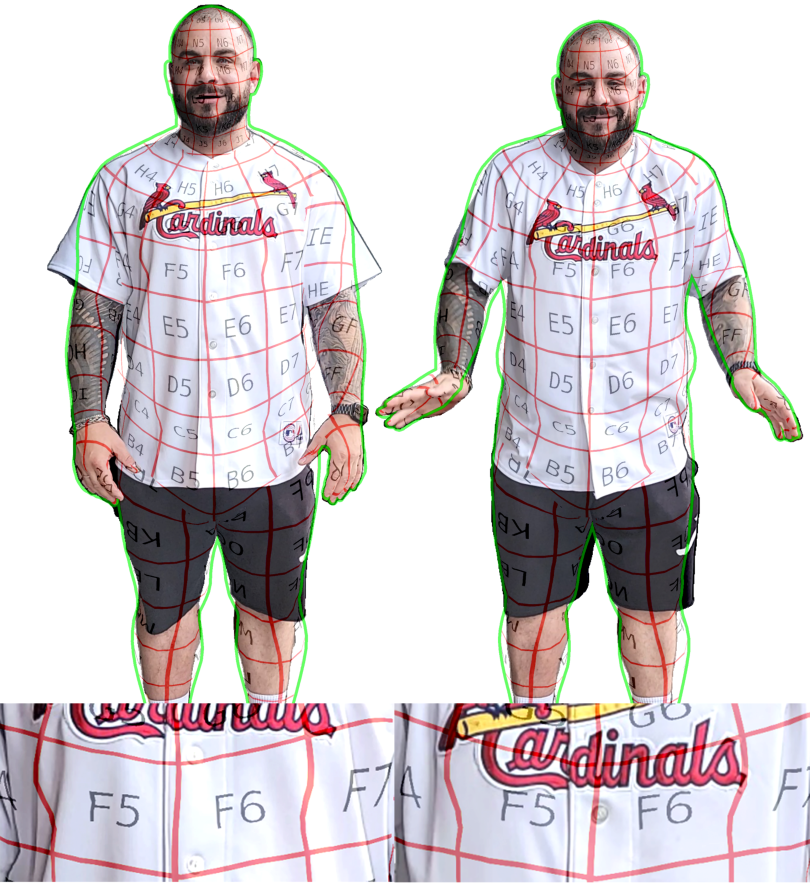}}
    \hspace{1em}
    \subfloat[\centering Texture space \label{fig:misalign:uv}]
    {\includegraphics[height=6.0cm]{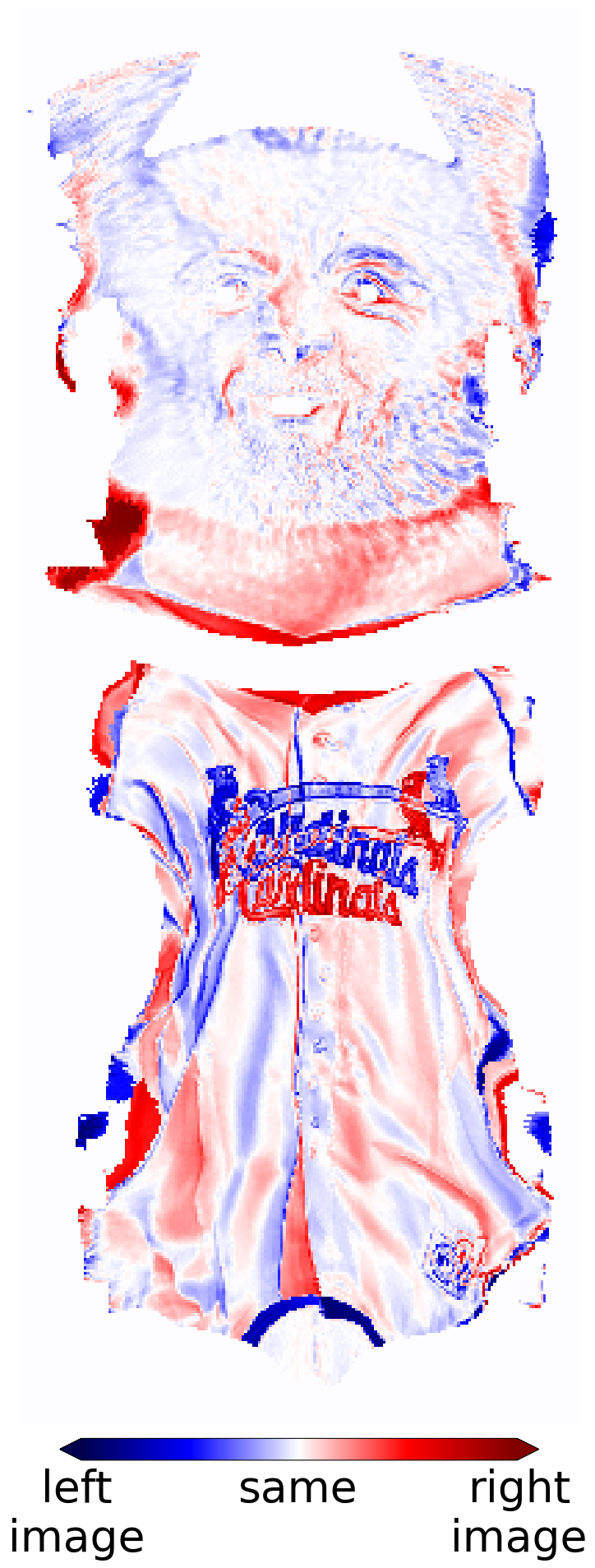}}%
\caption{\protect\subref{fig:misalign:rgb}~Mesh surfaces (represented by red lines) fitted to two different frames are located inconsistently.  Because of the rendering network's overfitting tendency, such misalignment can cause artifacts of the learned avatar. \protect\subref{fig:misalign:uv}~A partial plot of $\Pi_{\bf T}(\mathbf{I}_2) - \Pi_{\bf T}(\mathbf{I}_1)$, where $\Pi_{\bf T}$ is the texture space projection. The red and blue colors correspond to the two poses from \protect\subref{fig:misalign:rgb}, whilst the white color indicates that the colors match. The texture space is clearly misaligned.} 
\label{fig:misalign}
\end{figure}
}

{%
\setlength{\belowcaptionskip}{-6pt}
\renewcommand*{\arraystretch}{4}
\begin{figure}[t]
    \centering
    \subfloat[\centering\label{fig:warp-effect:gt}]
    {\includegraphics[width=0.32\columnwidth]{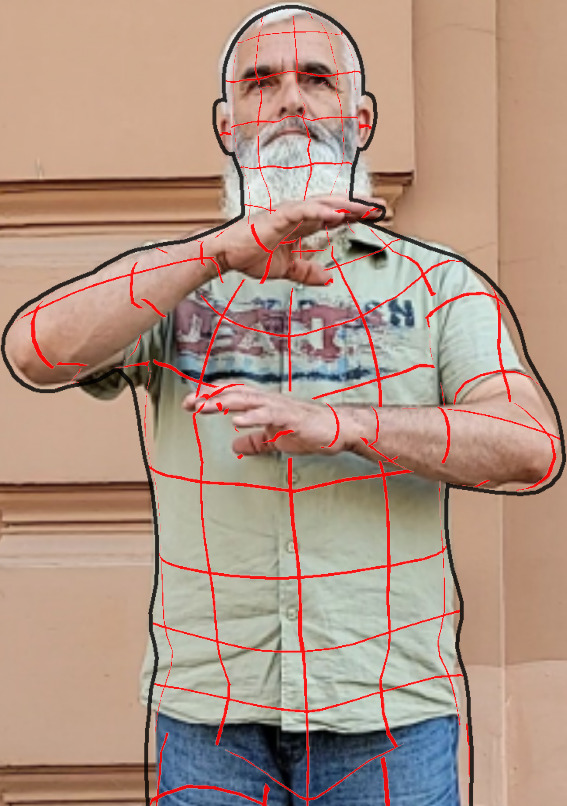}}%
    \subfloat[\centering\label{fig:warp-effect:warp-on}]
    {\includegraphics[width=0.32\columnwidth]{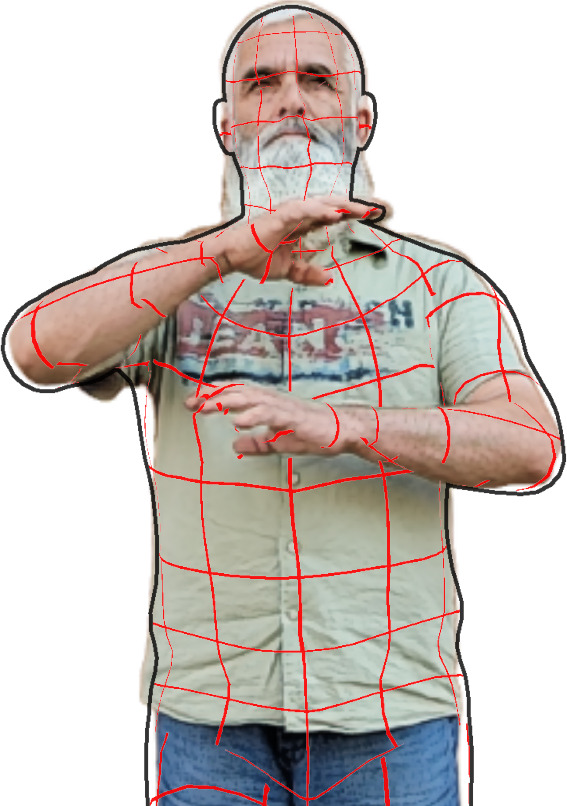}}%
    \subfloat[\centering\label{fig:warp-effect:warp-off}]
    {\includegraphics[width=0.32\columnwidth]{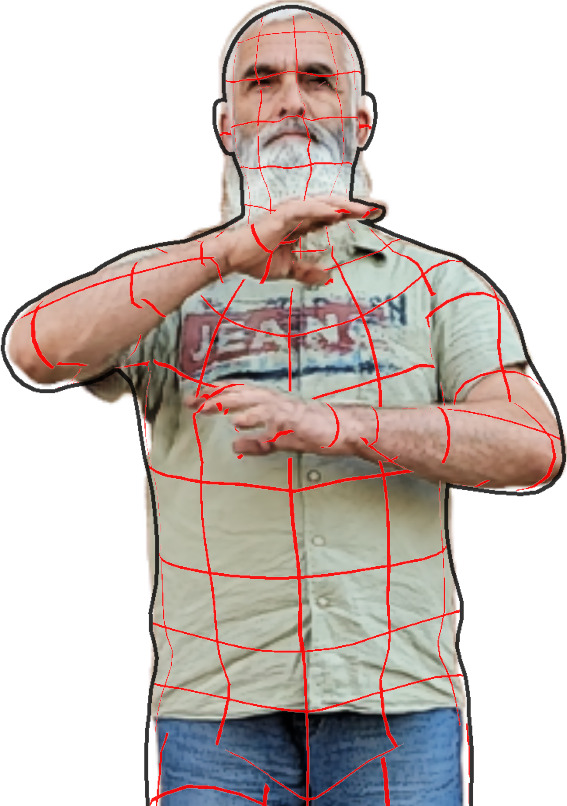}}%

\caption{\protect\subref{fig:warp-effect:gt} The \mbox{SMPL-X} (shown by red lines) superimposed on a training frame. \protect\subref{fig:warp-effect:warp-on} The corresponding \textbf{avatar} image produced by \titleabbreviation{}, that aligns with the training image well as compared to \protect\subref{fig:warp-effect:warp-off} where the learned texture warping field of the given frame was replaced with a zero warping field. Note, that the clothes appear in a \textit{neutral} position: the centralized shirt, the undeformed hemline. As seen, the warping field successfully ``absorbs'' the unique cloth deformation on this training frame, allowing the neural texture and the rendering network to learn an easier task.} %
\label{fig:warp-effect}
\end{figure}
}
The notable finding of \cite{dnr:2019} is that the quality of deferred neural rendering degrades very gracefully in the presence of coarseness or systematic errors in the geometric proxy. This has allowed several recent works~\cite{anr:2021,dynamic-garments:2021,dynamic-humans:2022,dressing-avatars:2022} to apply DNR to fullbody modeling using \mbox{SMPL-X} as an underlying geometry, despite the big gap between this geometric proxy and the actual avatar content that might include sliding or loose clothing. 
DNR based on \mbox{SMPL-X} geometry essentially ``off-loads'' clothing and hair modeling to the rendering network that operates in the 2D domain. While appealing for its simplicity,  we discovered that such an approach frequently fails to fit the training data, because frames with similar camera parameters and body pose can differ in hair geometry or clothing draping (see Figure~\ref{fig:misalign}), which the DNR approach cannot model well.  We propose modeling them in the texture space with 2D warping maps (see Figure~\ref{fig:arch}). Our cornerstone is that the texture space by construction is a canonical space for the human body -- it is invariant to body poses and camera views. At the same time it covers the whole 3D body surface. As a result, our warping approach is less ambiguous than commonly applied warping in the 3D spatial domain, and arguably is easier to train from monocular videos.

For each training frame, we learn a frame-specific warping field for the neural texture. The frame specific information is represented by a combination of the frame's \mbox{SMPL-X} pose vector~${\bf p}$ and a latent~$\bf z$. Those are passed through a multi-layered perceptron (MLP)~$\mathcal{W}$ to obtain a style vector~$\bf w$. We predict the warping fields with an encoder-decoder convolutional architecture~$\mathcal{E}\odot{}\mathcal{F}$, the input to which is a constant tensor of positional encodings of the texture coordinates. The style vector~$\bf w$ conditions the encoder part via the Adaptive Instance Normalization blocks~\cite{adain:2017}. A predicted warping field is a $2$-channel tensor, that specifies per-texel offsets of texture coordinates.  See supp. mat. for the exact details of the architectures $\mathcal{E}$, $\mathcal{F}$, $\mathcal{W}$. Figure~\ref{fig:warp-effect} shows the effect of the learned warping fields.

During model fitting, we use gradient descent to fit per-frame latent vectors $\bf z$, and parameters of the warping networks $\mathcal{E}$, $\mathcal{F}$, $\mathcal{W}$, the neural texture $\bf T$ and the rendering network $\mathcal{R}$, which are shared for all training frames. The latent vectors $\bf z$ encode the variations in avatar geometry that were not captured by the underlying \mbox{SMPL\nobreakdash-X} model, and the frame-specific imperfections of \mbox{SMPL\nobreakdash-X} fitting (Fig. \ref{fig:warp-effect}).

For rendering in novel poses, we compute the warping field corresponding to the neutral A-pose ${\bf p}$ and the average latent vector $\bf z$. Such warping field is applied to the neural texture once. First, such an approach is suitable in the case of limited train set.  Second, this helps eliminating the need for warping computations during inference, thus preserving high rendering speed.  %

\vspace{-3mm}
\paragraph{Texture initialization}\label{sec:method:texture-initialization}

 We found that, as compared to random initialization, better results can be achieved with \textit{spectral initialization} of the neural texture as in \mbox{StylePeople}~\cite{stylepeople}: each texel value is initialized with spectral coordinates of the corresponding point on the \mbox{SMPL-X} mesh graph (see examples in supp. mat.) We learn neural textures with small learning rates to avoid vanishing of the spectral information. We found that the use of spectral initialization was more pronounced in our system compared to \cite{stylepeople}. In particular, this initialization has minor effect on image fidelity, but it significantly slows down overfitting of the rendering network, giving the warping network time to converge and facilitating temporal consistency. It also contributes in plausible inpainting of body parts not covered in the input video (see Figure~\ref{fig:spectral}). 

{%
\setlength{\belowcaptionskip}{-6pt}
\begin{figure}
\centering
    \renewcommand*{\arraystretch}{4}
    \newcommand{\myfakeheight}{3.6cm}
    \subfloat[\centering\label{fig:spectral:05-rand}]
    {\includegraphics[trim=275 225 350 100,clip,height=\myfakeheight]{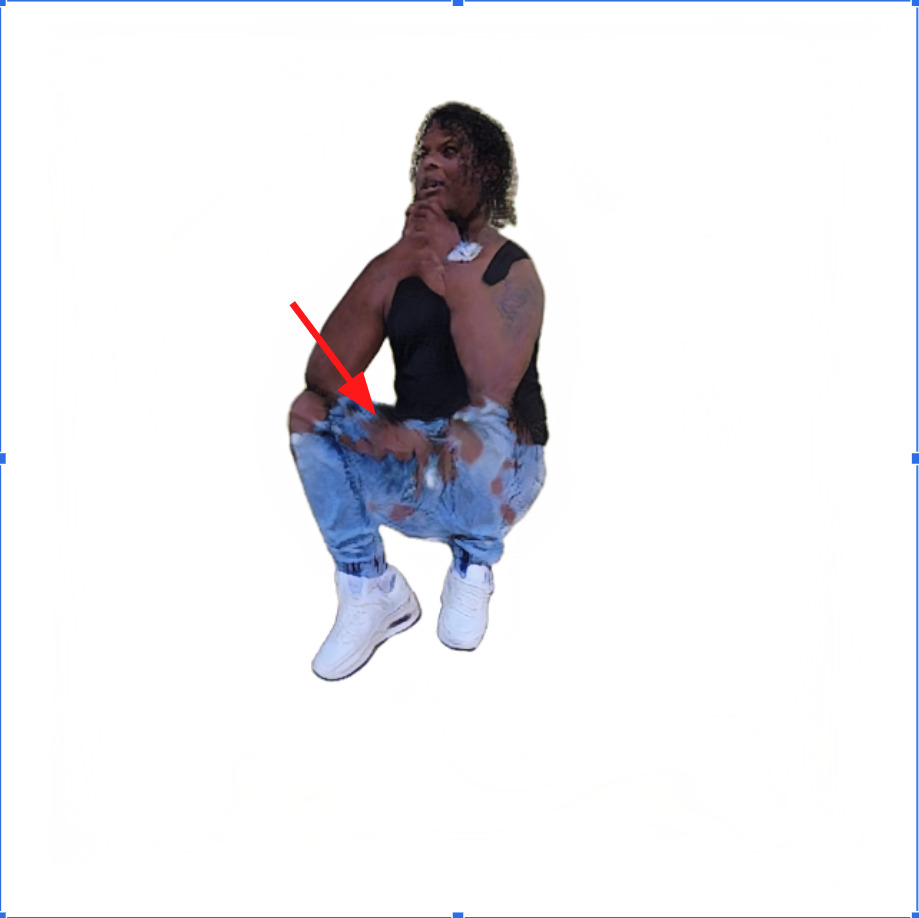}}
    \subfloat[\centering\label{fig:spectral:05-spectral}]
    {\includegraphics[trim=192 157 244 70,clip,height=\myfakeheight]{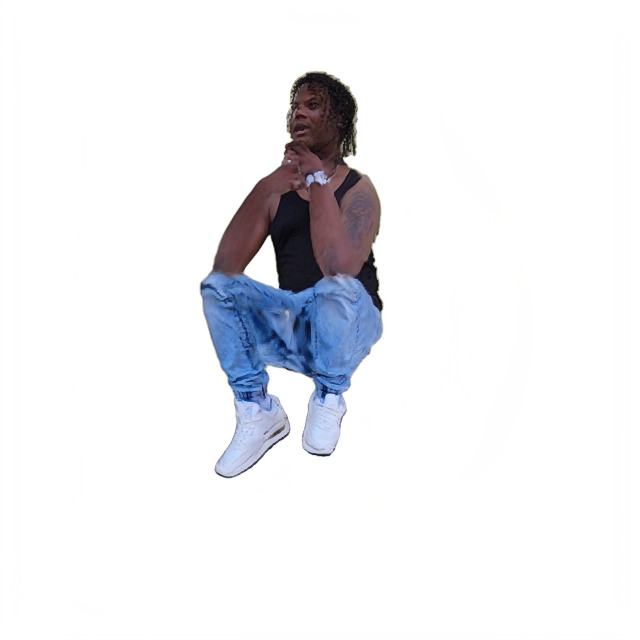}}%
    \hspace{1em}
    \subfloat[\centering\label{fig:spectral:08-rand}]
    {\includegraphics[height=\myfakeheight,trim={270 40 285 80},clip]{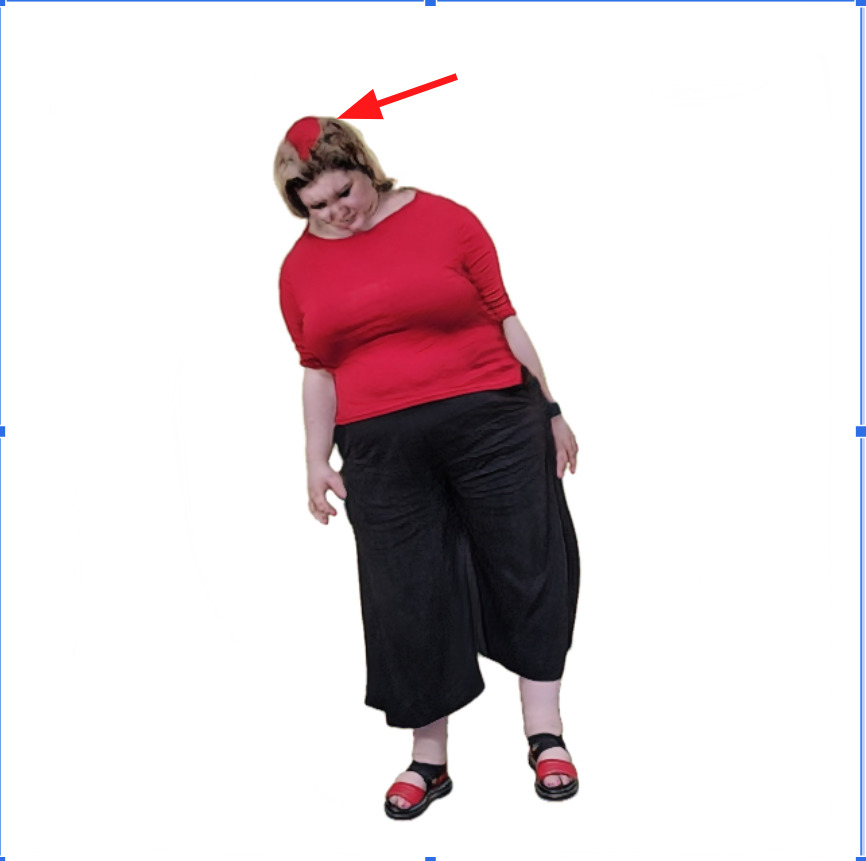}}%
    \subfloat[\centering\label{fig:spectral:08-spectral}]
    {\includegraphics[height=\myfakeheight,trim={200 30 211 60},clip]{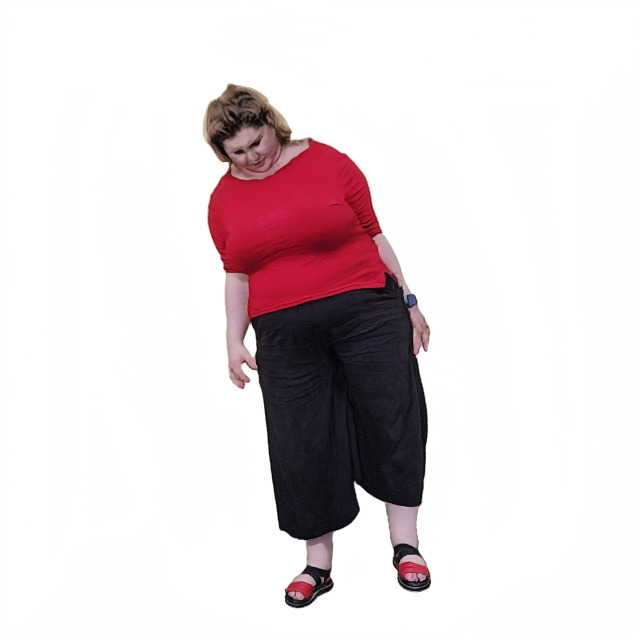}}%
\caption{Results of \titleabbreviation{} on novel poses, for two neural texture initialization methods: \protect\subref{fig:spectral:05-rand}/\protect\subref{fig:spectral:08-rand} - random initialization,  \protect\subref{fig:spectral:05-spectral}/\protect\subref{fig:spectral:08-spectral} - spectral initialization. Some body parts may exhibit overfitting artifacts. Local smoothness of spectral initialization acts as a prior to solve that.}
\label{fig:spectral}
\end{figure}
}

\section{Implementation details}

\subsection{Mesh fitting}\label{sec:details:meshfitting}

For both our system and the standard DNR model to perform well, precise and time-consistent \mbox{SMPL-X} mesh fitting to the training video frames (and at test time) is important. As for RGB photos, the convolutional regression-based techniques~\cite{vibe,hmr,expose,pixie,romp,bev,spin} produce somewhat accurate but inconsistent results that frequently mismatch the silhouettes. On the contrary, results of iterative non-linear optimization ~\cite{SMPLify:2016} better correspond 2D RGB images but are less plausible or unstable in 3D, since the fitting process relies on 2D image keypoints, which are poorly constrained along the depth dimension. Regression-based and optimization-based methods are unable to handle complex poses. Below we describe a few enhancements to \mbox{SMPL-X} optimization-based fitting that increase the stability of DNR-based avatar systems such as ours. We later validate the individual impact of these adjustments through ablation study.

\vspace{-4mm}
 \paragraph{Shared shape}
A person's identity does not change between frames of a training video, thus the shape-specific parameters of \mbox{SMPL-X} model can be shared.
\vspace{-4mm}
\paragraph{Silhouette loss}
We perform differentiable rendering of \mbox{SMPL-X} meshes via \mbox{\texttt{nvdiffrast}}~\cite{nvdiffrast}.
We then apply \mbox{Dice}~\cite{dice} loss to pull rendered \mbox{SMPL-X} silhouettes towards estimations of a semantic segmentation network.
In the presence of loose clothing, the inclusion of this term tends to ``fatten'' the human shape.

\vspace{-4mm}
\paragraph{Temporal loss}
The non-linear optimization-based fitting method tends to flicker on side views when the person's farther side is not visible. To compensate for this effect, we minimize the $\text{L2}$ distance between adjacent body pose vectors (in axis-angle format). This term is efficient for our scenario of training videos with slow body movements. Applying it to ``in-the-wild'' videos with fast motions may be ineffective or even harmful. %

\vspace{-4mm}
\paragraph{Automatic frame filtering}\label{sec:details:meshfitting:autofiltering}
The mesh fitting procedure is vulnerable to quick movements and depth ambiguity even with temporal guidance. Some training frames may be estimated with significantly incorrect poses. This causes a conflict between the 3D mesh surface and the neural texture, resulting in the appearance of ``ghost'' limbs and flickering. We use a simple heuristic that we describe in supp. mat. to automatically filter out bad frames (typically, less than 5\% of frames are discarded).

\subsection{Architectural details} 
\paragraph{Training} As a rendering network, we use U-Net~\cite{unet:2015} (a total of $\approx$14.4M parameters). Its decoder is based on bi-linear upsampling blocks. We provide the definitions of neural layers in supp. mat. We use ADAM~\cite{adam:2014} to optimize the architecture from scratch for a target person. On a single RTX 3090 GPU, convergence to plateau takes 8 hours. 
Our preliminary results suggest that faster acquisition is possible, if the architecture is pretrained on a dataset of people, and then fine-tuned to a target person. Below, however, we report on experiments that fit the model to a target video from scratch. %
\vspace{-4mm}
\paragraph{Ad-hoc inpainting of unseen body parts} Our minute-long training videos rarely include all body parts of the target person. Often top of the head, soles, and armpits have no ground truth. Those parts can be blurred on avatar renders, or worse -- excluded from the predicted segmentation. We overcome this by post-train finetuning for areas of the neural texture that correspond to unseen body parts. We first generate a pseudo ground-truth by projecting training frames into an RGB texture using \mbox{SMPL-X} fits, then we fill gaps on the RGB texture with nearby color, and render new ground-truth with extreme body and camera poses that show unseen body parts. We then fine-tune with L1 per-pixel loss. %
\vspace{-4mm}
\label{sec:details:priors}\paragraph{Inputs} We use \mbox{OpenPose}~\cite{OpenPose} to detect 2D keypoints, which is required for mesh fitting. We retrain \mbox{Graphonomy}~\cite{graphonomy:2019} segmentation network on \mbox{CCIHP} dataset~\cite{ccihp_dataset_2021}, which includes accessories like watches, belts, and glasses, etc. Meanwhile the publicly available \mbox{Graphonomy} trained on \mbox{CIHP} dataset predicts those as background.

\subsection{Loss functions}\label{sec:method:loss-functions}
For \titleabbreviation{} training, we use popular loss functions from previous works~\cite{dnr:2019,stylepeople,pix2pix} with modifications. We further elaborate on our learning objective:
{\setlength{\abovedisplayskip}{0.5em}
\setlength{\abovedisplayshortskip}{0.5em}
\setlength{\belowdisplayshortskip}{0.5em}
\setlength{\belowdisplayskip}{0.5em}
\label{eq:total_loss}
\begin{equation}
\begin{gathered}
\mathcal{L}_{\text{total}} = \sum_{i \in \mathfrak{L}} \lambda_i \mathcal{L}_i \\
\mathfrak{L} = \{\text{L1}, \text{Percept}, \text{GAN}, \text{Dice},\text{TV}_\text{RGB}, \text{TV}_\Delta, \Delta\shortrightarrow0 \}
\end{gathered}
\end{equation}
}
\vspace{-4mm}
\paragraph{Image losses} These include $\text{L1}$ per-pixel loss $\mathcal{L}_{\text{L1}}$ and the VGG \cite{vgg} based perceptual loss~\cite{percept} $\mathcal{L}_{\text{Percept}}$. To enforce the realism of the rendered images, we utilized the non-saturating adversarial loss~\cite{nsgan} $\mathcal{L}_{\text{GAN}}$ on predictions of multi-scale patch-based discriminators (\mbox{PatchGAN} from \mbox{Isola \etal}~\cite{patch-gan}). 

\vspace{-4mm}
\paragraph{Segmentation loss} In our case, the estimated body mesh (\mbox{SMPL-X}) is coarse and frequently misaligned with the underlying human content. To assist the renderer in generating content outside of the rasterized mesh, the renderer predicts the human mask in addition to the RGB image. We supervise the masks on the pseudo ground truth masks (see paragraph \ref{sec:details:priors}), using Dice~\cite{dice} $\mathcal{L}_{\text{Dice}}$ segmentation loss.
\vspace{-4mm}
\paragraph{Regularization} For the warping fields, we penalize $\text{L2}$ norm ($\mathcal{L}_{\Delta\shortrightarrow0}$) to prevent large deviations from identity warping, and also apply total-variation (TV) loss~\cite{tv-functional:2004} $\mathcal{L}_{\text{TV}_\Delta}$ for local smoothness. We use a low weighted TV loss on generated images $\mathcal{L}_{\text{TV}_{\text{RGB}}}$ to slightly improve smoothness.

\subsection{Avatar rendering on a mobile device}
We developed a telepresence mobile app for Qualcomm Snapdragon SoC using \mbox{Android}, \mbox{ARCore}~\cite{arcore}, and \mbox{OpenGL}. Our method can run natively at resolution of $640\times640$ pixels and above 30 frames per second (measured on Qualcomm Snapdragon 888, see Fig.~\ref{fig:smartphone}). We use \mbox{OpenGL} shaders to generate input neural rasterizations and to infer posed \mbox{SMPL-X} meshes from body joint rotations. We stumbled upon \mbox{OpenGL} limit on the depth of the neural texture channels. We were able to pack at most 16 channels of neural rasterizations into \mbox{OpenGL} framebuffers, using quantization \cite{quantization-whitepaper:2021} into eight-bit integers. Similarly, we use post-training quantization on the weights of the rendering network. This enables the network to run at high speeds on the Qualcomm Digital Signal Processor. Quality degradation due to quantization is relatively low in our system. Because single-person avatars have low variance in appearance, the numerical range of neural network activations is small, which is advantageous for quantization.

{%
\setlength{\belowcaptionskip}{0pt plus 2pt minus 2pt}
\begin{figure*}[t]
\centering
    \newcommand{\myfakeheight}{4.8cm}
    \newcommand{\myfakeraise}{2.3cm}
    \newcommand{\myfakeheightzoom}{0.97cm}
    \newcolumntype{P}[1]{>{\centering\arraybackslash}p{#1}}
    \setlength{\tabcolsep}{0pt} %
    \begin{tabular}{P{1.55cm}P{1.8cm}P{1.8cm}P{1.75cm}P{1.5cm}P{1.55cm}P{1.75cm}P{1.75cm}l} %
    \includegraphics[height=\myfakeheight,trim={300 70 400 70},clip]{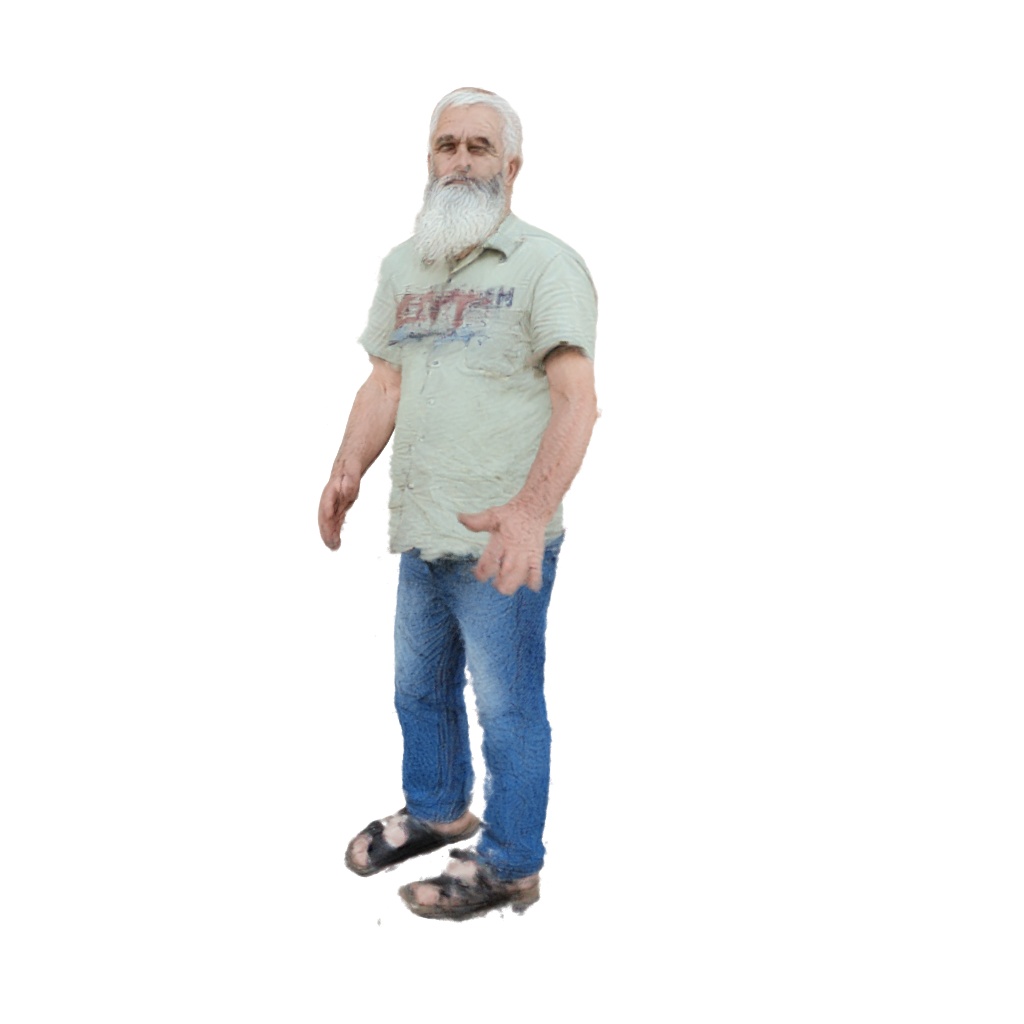}
    & \includegraphics[height=\myfakeheight,trim={345 165 370 85},clip]{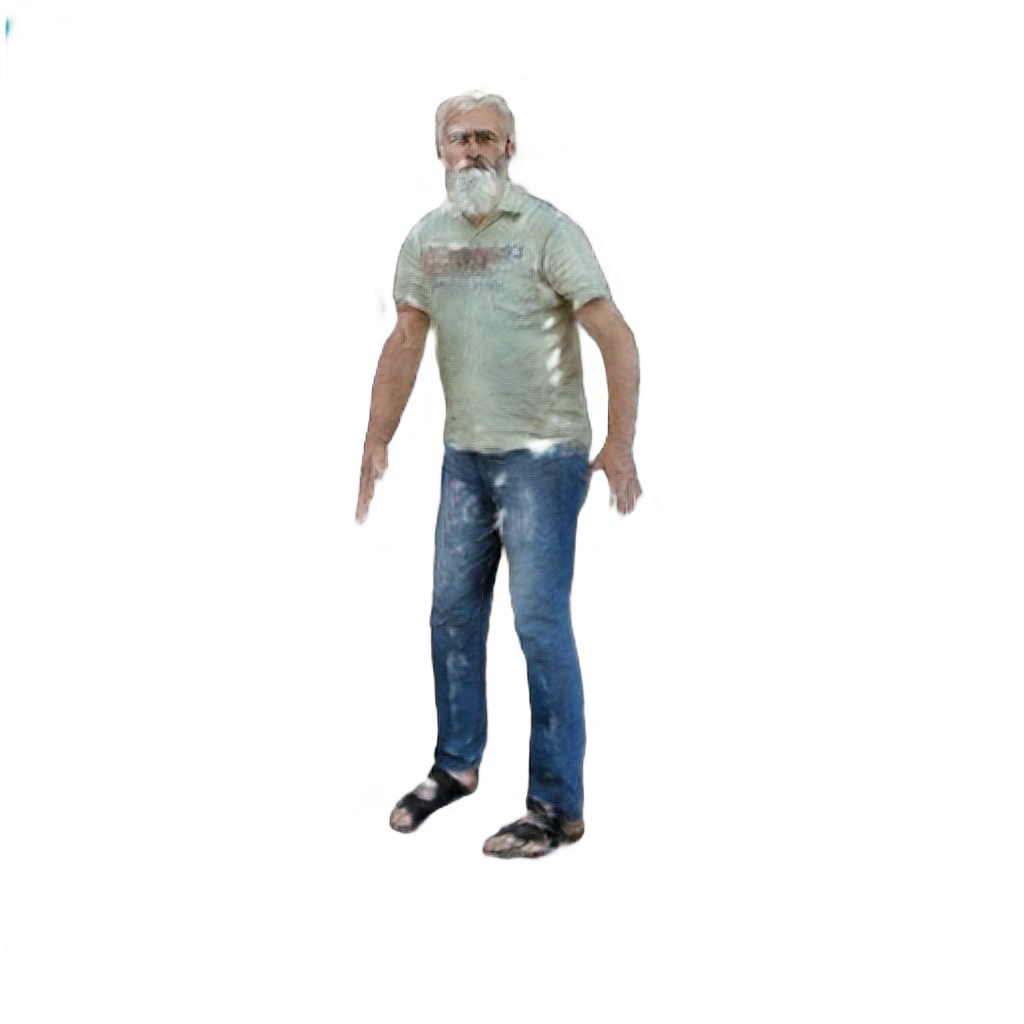}
    & \includegraphics[height=\myfakeheight,trim={150 20 190 20},clip]{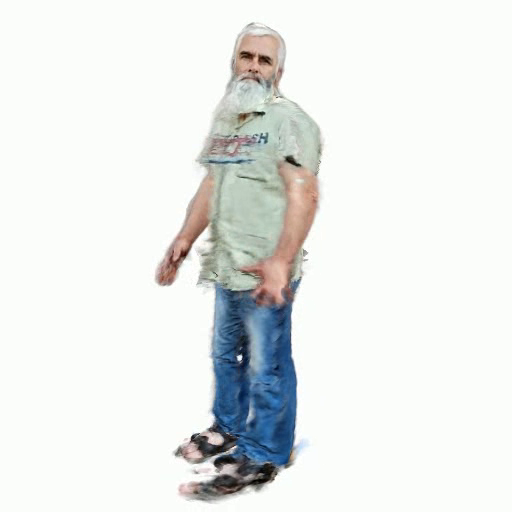}
    & \includegraphics[height=\myfakeheight,trim={215 20 200 20},clip]{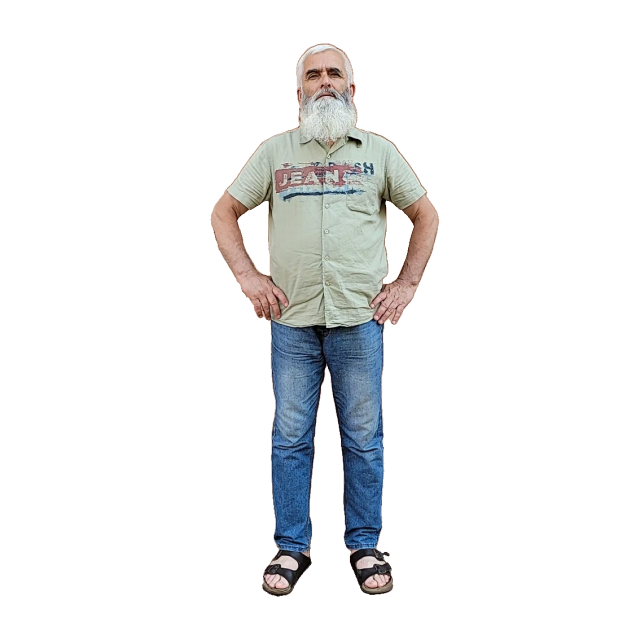}
    & \includegraphics[height=\myfakeheight,trim={200 20 230 20},clip]{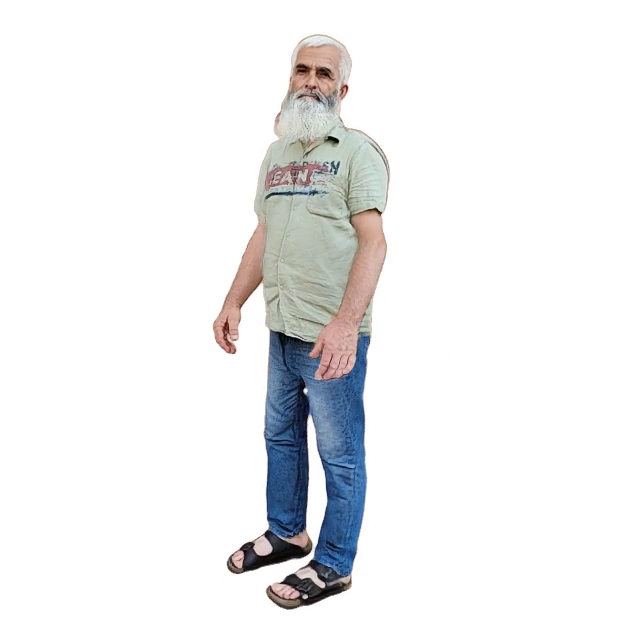}
    & \includegraphics[height=\myfakeheight,trim={200 20 230 20},clip]{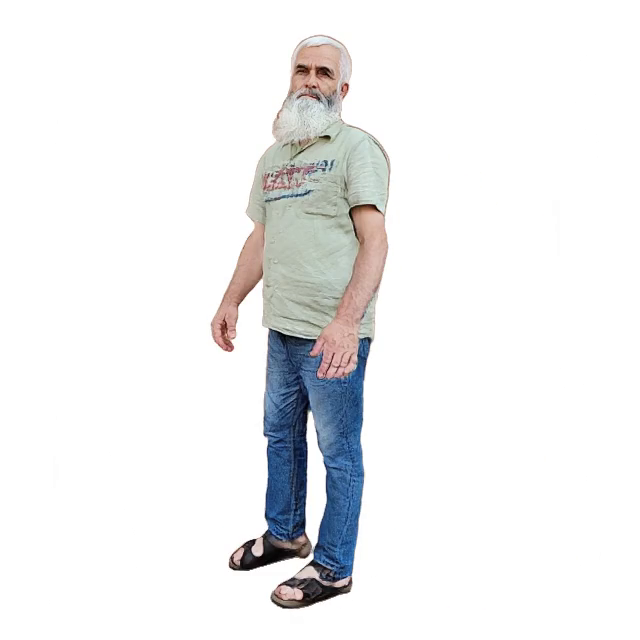}
    &  \includegraphics[height=\myfakeheight,trim={210 20 250 20},clip]{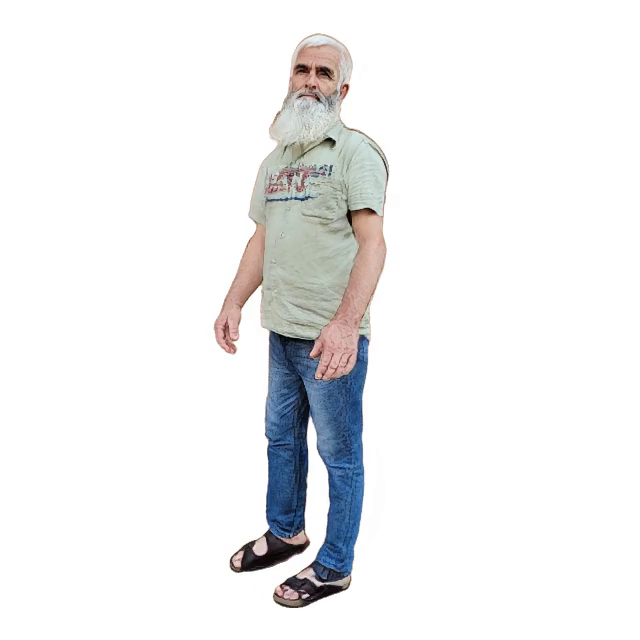}
    & \includegraphics[height=\myfakeheight,trim={165 25 160 25},clip]{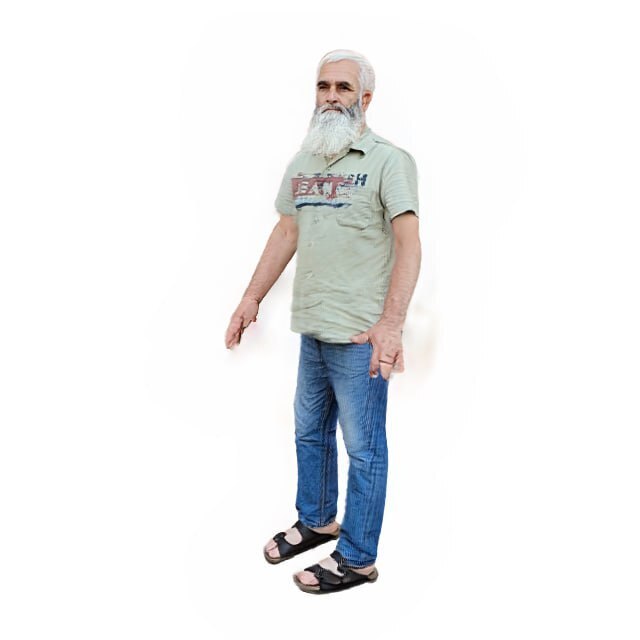}
    & \raisebox{\myfakeraise}{\makecell[l]{
    \raisebox{-.5\height}{\includegraphics[height=\myfakeheightzoom,trim={230 420 230 150},clip]{img/methods-compare/gt/02.png}}\ Target\\
    \raisebox{-.5\height}{\includegraphics[height=\myfakeheightzoom,trim={230 420 230 150},clip]{img/methods-compare/warp/403_seq_02.png}}\ \underline{\textbf{\titleabbreviation{}}}\\
    \raisebox{-.5\height}{\includegraphics[height=\myfakeheightzoom,trim={230 420 230 150},clip]{img/methods-compare/anr-r-anr/403_seq_02.png}}\ ANR\\
    \raisebox{-.5\height}{\includegraphics[height=\myfakeheightzoom,trim={230 420 230 150},clip]{img/methods-compare/dnr/403_seq_02.png}}\ \makecell[l]{Style\\People}\\
    \raisebox{-.5\height}{\includegraphics[height=\myfakeheightzoom]{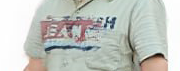}}\ \makecell[l]{Human\\NeRF}}}
    \\ 
    \includegraphics[height=\myfakeheight,trim={345 90 448 130},clip]{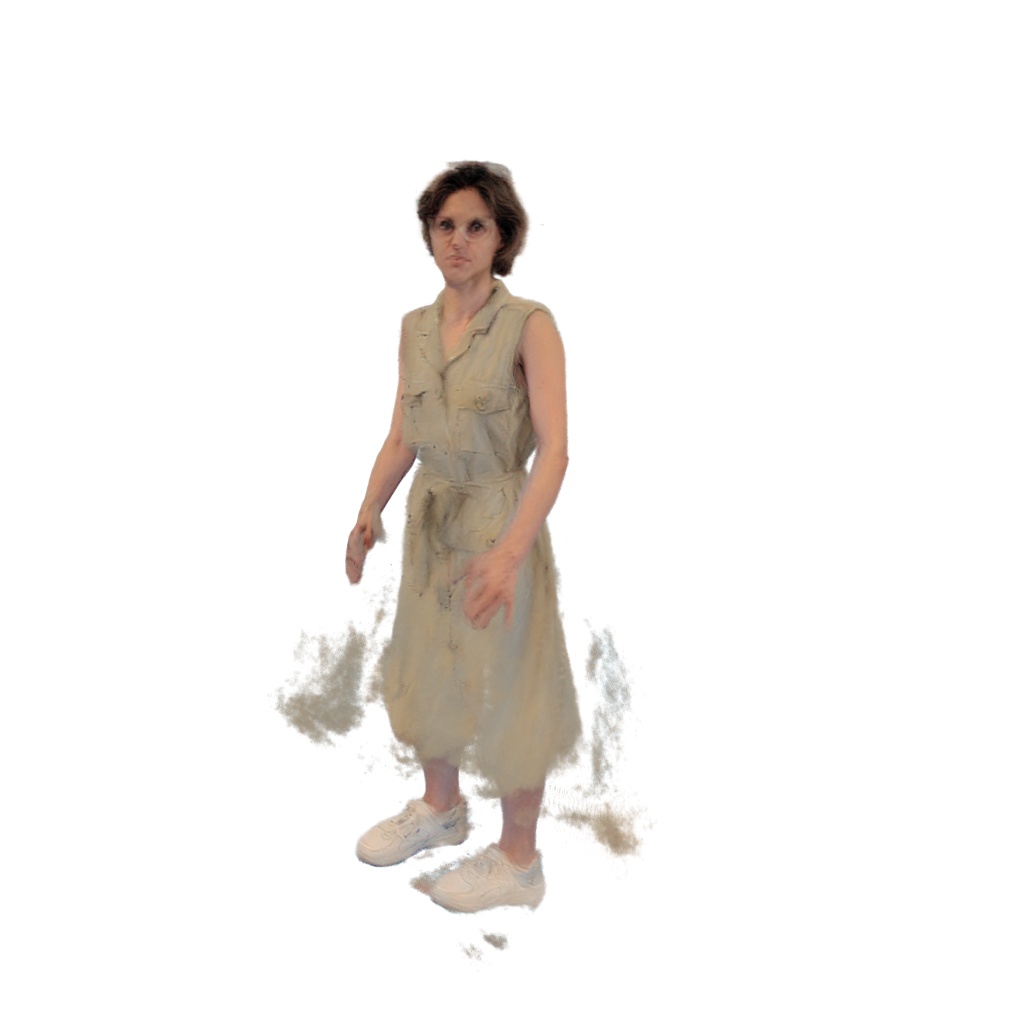}
    & \includegraphics[height=\myfakeheight,trim={345 115 370 85},clip]{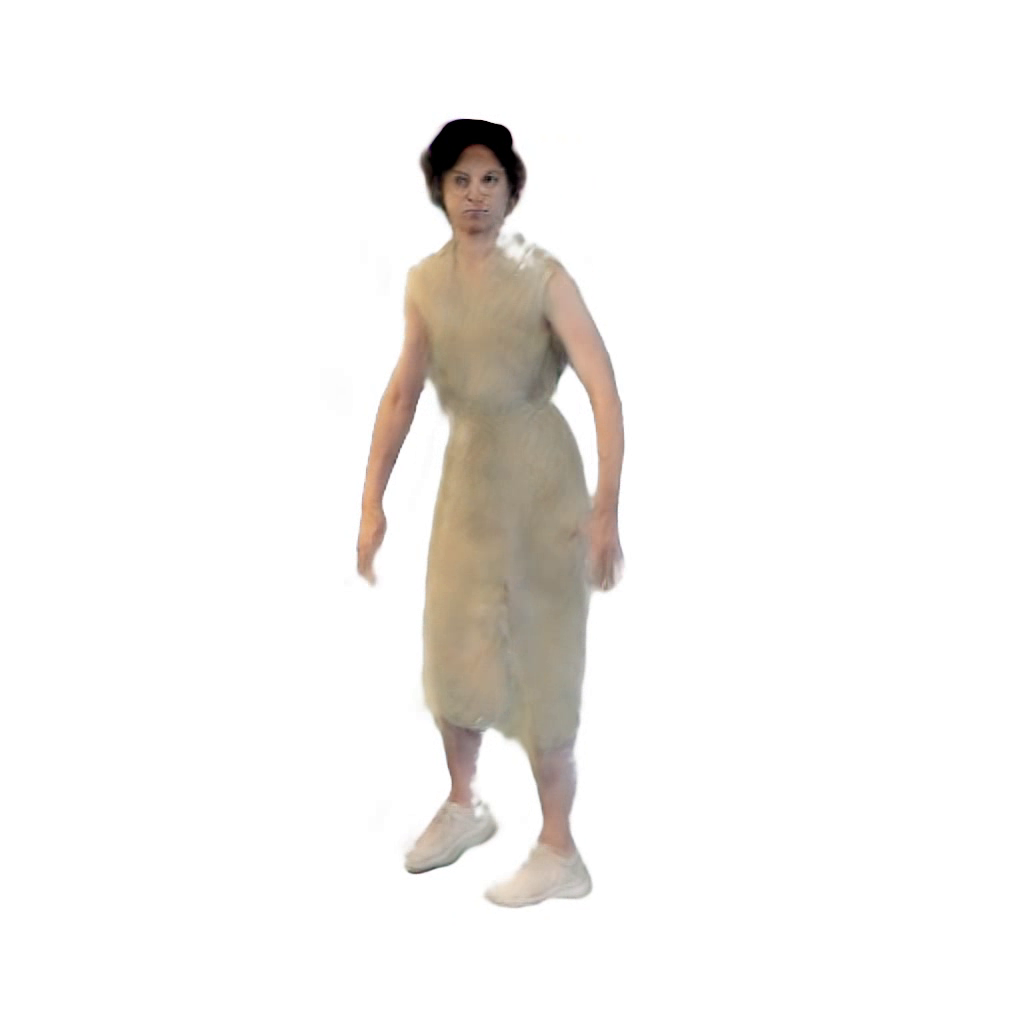}
    & \includegraphics[height=\myfakeheight,trim={150 0 190 10},clip]{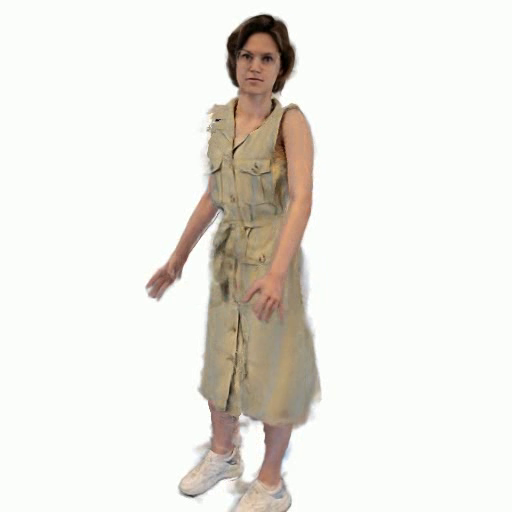}
    & \includegraphics[height=\myfakeheight,trim={240 40 230 40},clip]{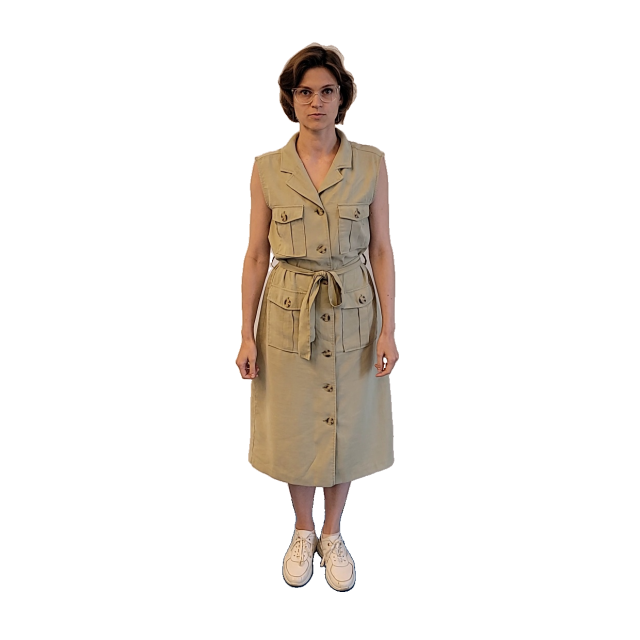}
    & \includegraphics[height=\myfakeheight,trim={200 20 230 20},clip]{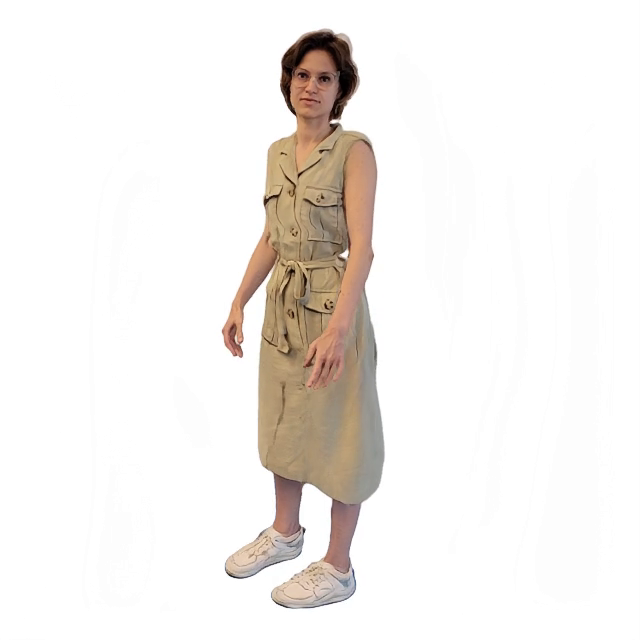}
    & \includegraphics[height=\myfakeheight,trim={200 20 230 20},clip]{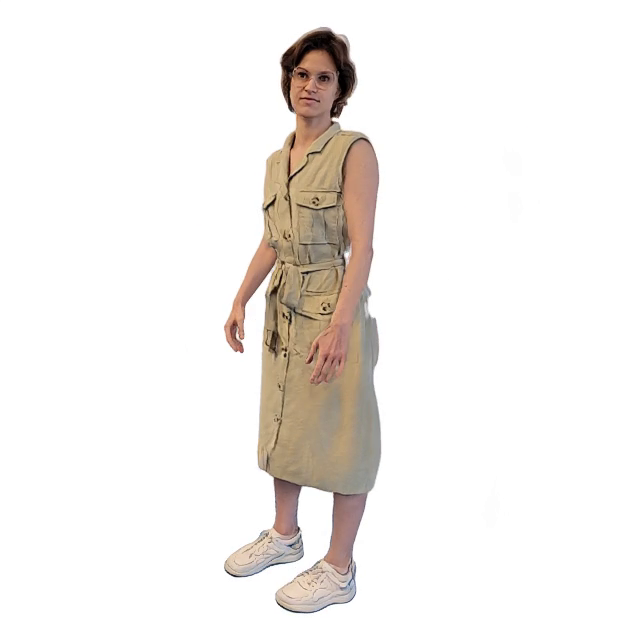}
    & \includegraphics[height=\myfakeheight,trim={210 20 250 20},clip]{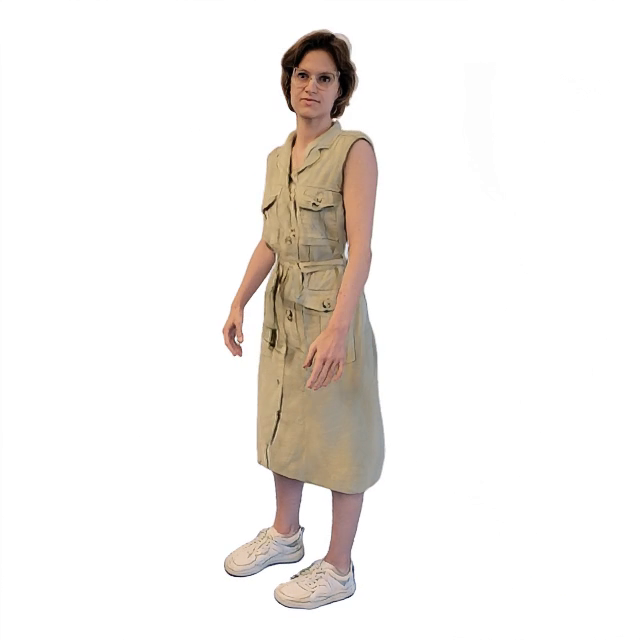}
    & \includegraphics[height=\myfakeheight,trim={165 25 170 25},clip]{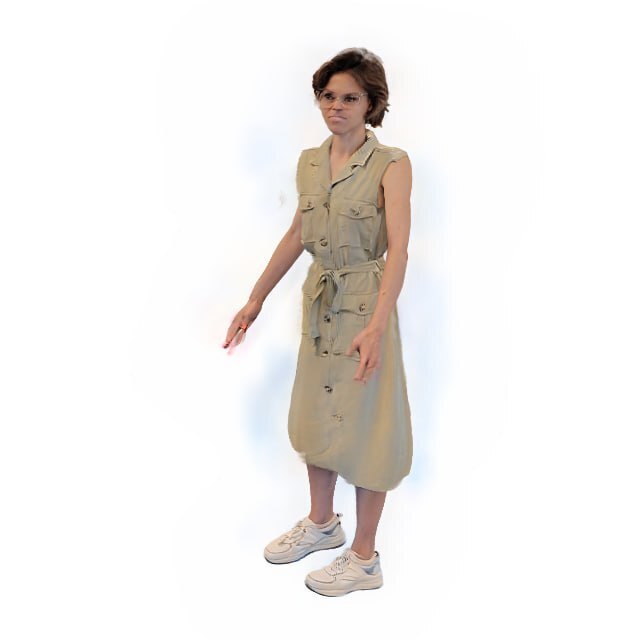}
    & \raisebox{\myfakeraise}{\makecell[l]{
    \raisebox{-.5\height}{\includegraphics[height=\myfakeheightzoom,trim={230 310 230 260},clip]{img/methods-compare/gt/09.png}}\ Target\\
    \raisebox{-.5\height}{\includegraphics[height=\myfakeheightzoom,trim={230 320 230 250},clip]{img/methods-compare/warp/403_seq_09.png}}\ \underline{\textbf{\titleabbreviation{}}}\\
    \raisebox{-.5\height}{\includegraphics[height=\myfakeheightzoom,trim={230 320 230 250},clip]{img/methods-compare/anr-r-anr/403_seq_09.png}}\ ANR\\
    \raisebox{-.5\height}{\includegraphics[height=\myfakeheightzoom,trim={230 320 230 250},clip]{img/methods-compare/dnr/403_seq_09.png}}\ \makecell[l]{Style\\People}\\
    \raisebox{-.5\height}{\includegraphics[height=\myfakeheightzoom]{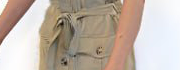}}\ \makecell[l]{Human\\NeRF}}}
    \\ 
    \multicolumn{1}{c}{\small InstantAvatar } %
    & \multicolumn{1}{c}{\small HF-Avatar} %
    & \multicolumn{1}{c}{\small AnimNeRF} %
    & \multicolumn{1}{|c|}{\small Target}
    & \multicolumn{1}{c}{\small \underline{\textbf{\titleabbreviation{}}}}
    & \multicolumn{1}{c}{\small ANR}
    & \multicolumn{1}{c}{\small StylePeople} %
    & \multicolumn{1}{c}{\small HumanNeRF } %
    & \multicolumn{1}{l}{\kern1cm \small \textit{(zoom in)}}
    \\
    \multicolumn{3}{c}{\small $512\times512$ px, on A-poses}
    & \multicolumn{1}{|c|}{\small identity}
    & \multicolumn{5}{c}{\small $640\times640$ px, on full training sequences}
    \\
    \cline{1-3}\cline{5-9}
    \end{tabular}
\caption{Avatars created with different monocular video-based approaches. \titleabbreviation{} achieves the highest visual quality in this comparison. See text for more details.} 
\label{fig:methods-comparison}
\end{figure*}
}

\section{Experiments}\label{sec:experim}

We compare our method to similar recent approaches and provide an ablation study for our mesh fitting adjustments. We conduct quantitative comparisons as well as user study comparisons. We provide sample side-by-side comparisons of the methods in supp. mat.

\subsection{Self-captured dataset}
We have gathered a dataset of stationary monocular videos of ten people with different demographics, body shapes, and clothes topology. We note that despite improvements in mesh fitting, it is still imperfect, therefore we captured the training videos with a specific scenario that avoids complex poses. In one-minute training videos people move slowly, turn $360^\circ$, and show their hands in front of their torso. The one-minute hold-out videos include mostly novel poses not present in training videos, namely various arm and hip movements. See samples of training and hold-out data in supp. mat. %

\begingroup
\renewcommand*{\arraystretch}{4}

\begin{figure*}[t]
    \centering
    \setlength{\abovecaptionskip}{0pt plus 2pt minus 2pt}
    \setlength{\belowcaptionskip}{0pt plus 2pt minus 2pt}
    \newcommand\rowincludegraphics[2][]{\raisebox{-0.45\height}{\includegraphics[#1]{#2}}}
    \newcommand{\myfigheight}{1.42cm}
    \newcommand{\myparboxwidth}{0.5cm}
    \newcommand{\myparboxsep}{1.5em}
    \setlength{\tabcolsep}{0pt} %
    \setlength{\arrayrulewidth}{1pt}

    \begin{tabular}{m{1em}Slp{1em}SlSlm{1em}}
    \parbox[c]{\myparboxwidth}{\rotatebox[origin=c]{90}{\titleabbreviation{} \hspace{\myparboxsep} SP}}
    & \makecell{
    \rowincludegraphics[height=\myfigheight]{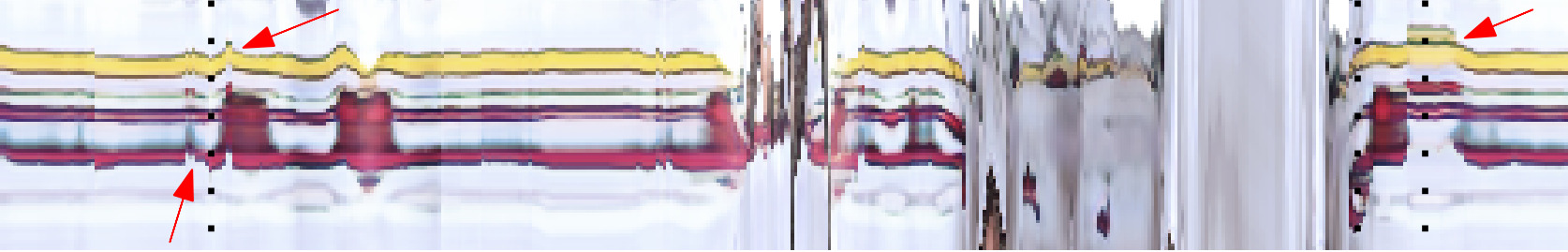} 
    \\ \rowincludegraphics[height=\myfigheight]{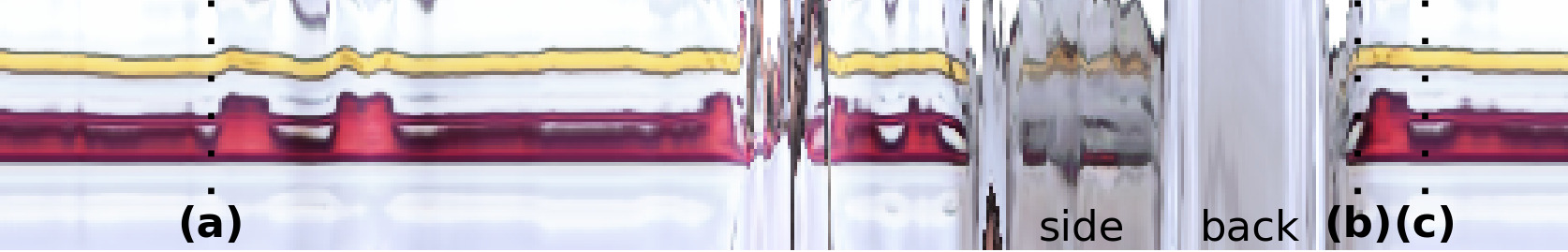}} &
    & \makecell{
    \rowincludegraphics[height=\myfigheight]{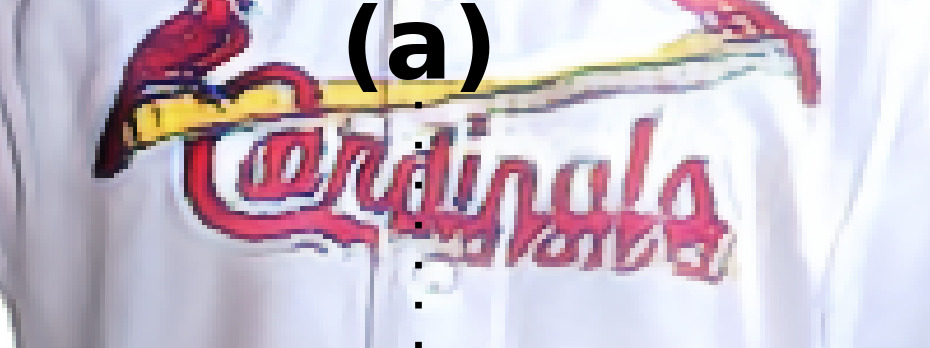} \\ \rowincludegraphics[height=\myfigheight]{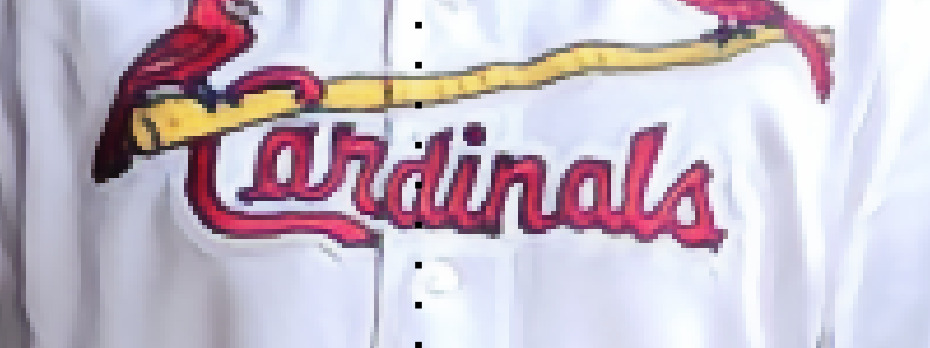}} 
    & \makecell{
    \rowincludegraphics[height=\myfigheight]{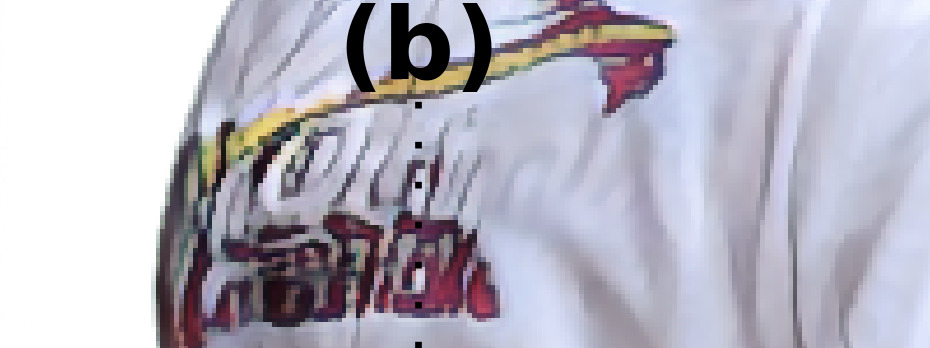} \\
    \rowincludegraphics[height=\myfigheight]{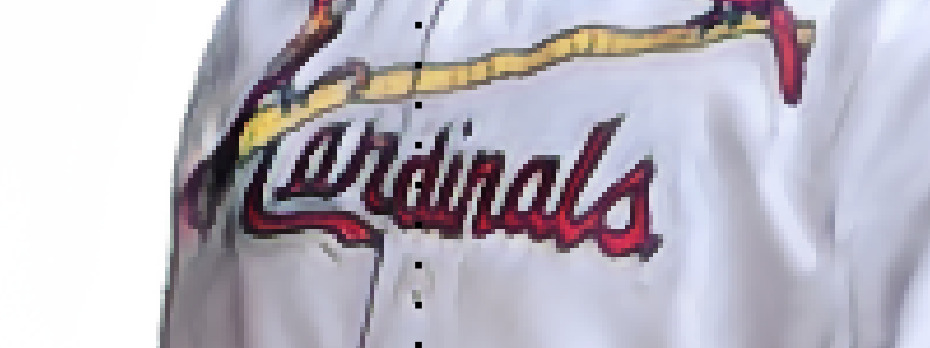}}
    & \parbox[c]{\myparboxwidth}{\rotatebox[origin=c]{90}{\titleabbreviation{} \hspace{\myparboxsep} SP}} \\\cline{1-4}
    \parbox[c]{\myparboxwidth}{\rotatebox[origin=c]{90}{\titleabbreviation{} \hspace{\myparboxsep} SP}}
    & \makecell{
    \rowincludegraphics[height=\myfigheight]{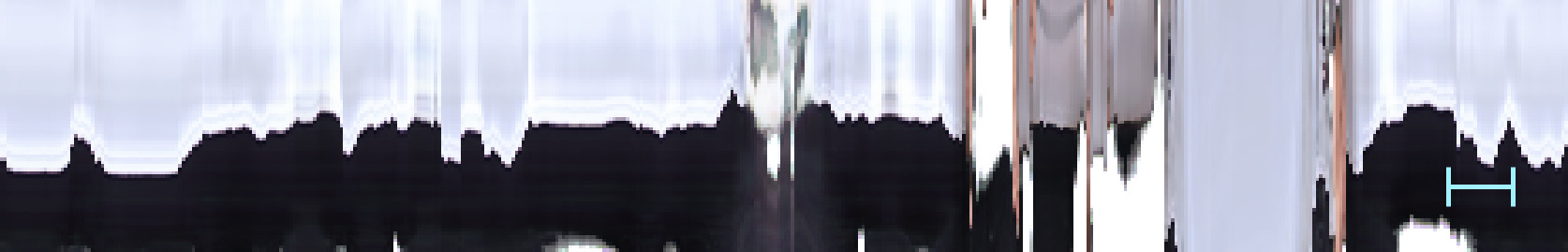}\\
    \rowincludegraphics[height=\myfigheight]{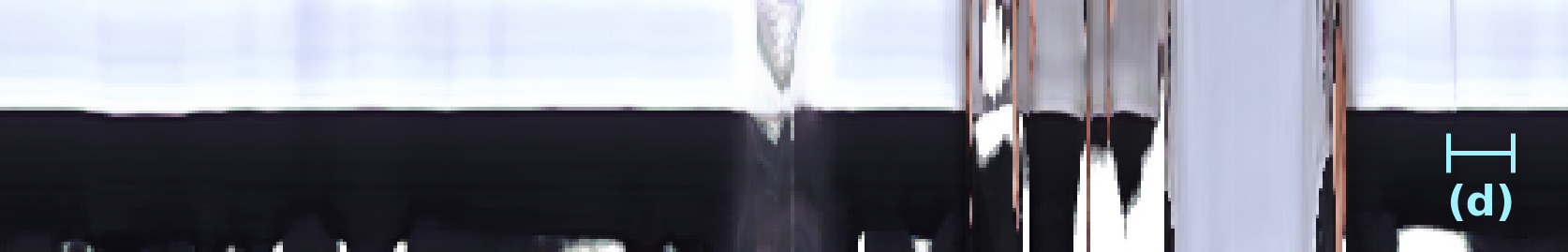}} &
    & \makecell{
    \rowincludegraphics[height=\myfigheight]{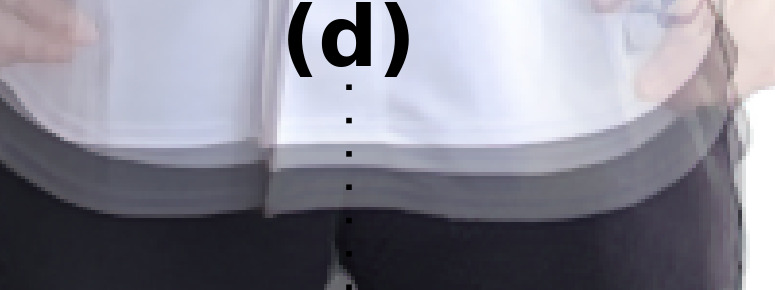} \\ \rowincludegraphics[height=\myfigheight]{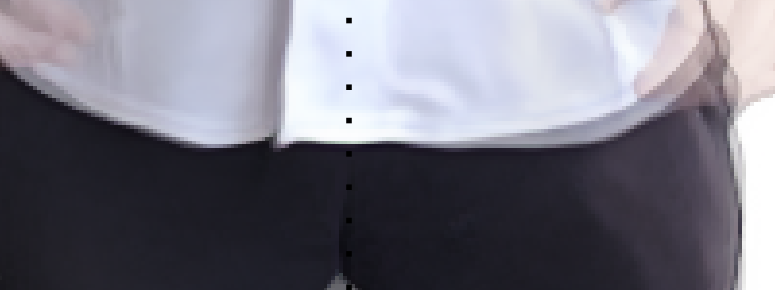}}
    & \multicolumn{1}{|c}{\makecell{
    \rowincludegraphics[height=\myfigheight]{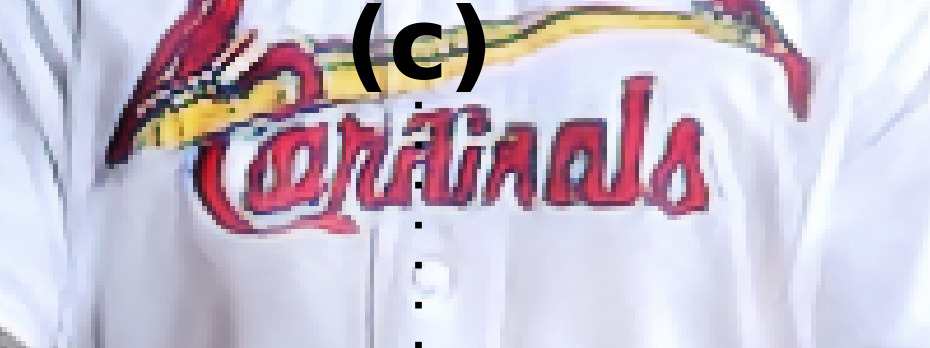}\\
    \rowincludegraphics[height=\myfigheight]{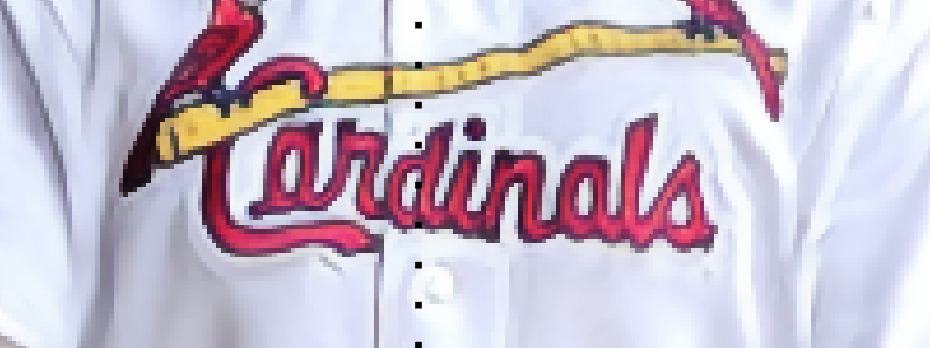}}}
    & \parbox[c]{\myparboxwidth}{\rotatebox[origin=c]{90}{\titleabbreviation{} \hspace{\myparboxsep} SP}} \\ \cline{5-5}
    \end{tabular}

    \subfloat{\label{fig:dnr-vs-warp:a-tshirt}}
    \subfloat{\label{fig:dnr-vs-warp:b-tshirt}}
    \subfloat{\label{fig:dnr-vs-warp:c-tshirt}}
    \subfloat{\label{fig:dnr-vs-warp:d-hem}}

    \caption{\label{fig:dnr-vs-warp} To illustrate the difference in temporal consistency, we show the space-time $y$-$t$ slices (left). In the case of \mbox{StylePeople} (SP), the T-shirt pattern (top plots) abruptly tears vertically (accentuated with red arrows), resulting in \textbf{(a)} flickering,  \textbf{(b)} ``ghost'' patterns, \textbf{(c)} poor clarity, and loss of pattern rendering for rotated human body (see the section labeled \textit{``side''}). These artifacts are absent in our approach's results, which also achieves good temporal stability on the hemline (bottom plots). \textbf{(d)} Shows an average image over a 20-frame interval (colored in light-blue).}  
    
\end{figure*}
\endgroup

\vspace{-4mm}
{%
\setlength{\belowcaptionskip}{0pt plus 2pt minus 2pt}
\setlength{\tabcolsep}{4pt}

\newcommand{\myup}[1][black]{\textcolor{#1}{\ensuremath\uparrow}}
\definecolor{olivegreen}{HTML}{00A64F}
\begin{table}[!h]
\renewcommand*{\arraystretch}{0.9}
\caption{Quantitative metrics and the results of the user preference study, in which each method is compared against \titleabbreviation{}. See the text for the discussion of the results.%
}
\label{tab:metrics}
\centering
\begin{tabular}{c|ccc||c} 
 \hline
 & SSIM   & LPIPS        & tLPIPS             & {\small \% of users} \\
 & $\cdot10^2$↑ & $\cdot10^2$↓ & $\cdot10^3$↓   & {\small preferred \textbf{\titleabbreviation{}}} \\
 \hline
 \multicolumn{5}{c}{\textit{Scenario A: $512\times512$ px, training only on A-poses}} \\
 \hline
 InstantAvatar & 92.52 & 7.91 & 7.52 & 96.5\% {\myup[olivegreen]} \\ 
 HF-Avatar & 93.0 & 7.25 & 4.00 & 90.0\% {\myup[olivegreen]} \\ 
 AnimNeRF & 92.7 & 6.74 & 2.96{} & 98.0\% {\myup[olivegreen]}  \\ 
 \textbf{\titleabbreviation{}} & \cellcolor{green!25}{94.5} & \cellcolor{green!25}{3.95} & \cellcolor{green!25}{2.56} & N/A \\
 \hline
 \multicolumn{5}{c}{\textit{Scenario B: $640\times640$ px, full training sequences}} \\
 \hline
 HumanNeRF & 87.4 & 13.00 & 19.3 & 95.0\% {\myup[olivegreen]}\\
 StylePeople & \cellcolor{green!25}{90.0}  & \cellcolor{green!25}{6.13} & 4.77 & 60.5\% {\myup[olivegreen]}\\
 ANR & \cellcolor{green!25}{90.0}  & \cellcolor{yellow!25}{6.14} & \cellcolor{yellow!25}{4.76} & 69.2\% {\myup[olivegreen]} \\
 \textbf{\titleabbreviation{}} & \cellcolor{yellow!25}{89.9}  & 6.23{} & \cellcolor{green!25}{4.66} & N/A\\
 \hline
\end{tabular}
\end{table}
}

\begin{table}[!h]
\renewcommand*{\arraystretch}{0.9}
\caption{Mesh fitting used in training and validation.}
\label{tab:mesh-fitting}
\centering
\rowcolors{1}{gray!10}{white}
\begin{tabular}{c|c|c} 
 \hline
 & \textbf{Training} & \textbf{Hold-out} \\
\hline
 InstantAvatar & their SMPL & their SMPL \\
 HF-Avatar & their SMPL & \makecell{our SMPL \& \\ their train body shape} \\
 AnimNeRF & their SMPL & their SMPL \\
 HumanNeRF & our SMPL & our SMPL \\
 StylePeople & our SMPL-X & our SMPL-X\\
 ANR & our SMPL-X &  our SMPL-X\\
 \textbf{\titleabbreviation{}} &  our SMPL-X &  our SMPL-X \\
 \hline
\end{tabular}
\end{table}

We use publicly available implementations of \mbox{InstantAvatar}, \mbox{HF-Avatar}, \mbox{Anim-NeRF}, \mbox{NeuMan}  and \mbox{HumanNeRF}. To compare with \mbox{InstantAvatar}, \mbox{HF-Avatar}, \mbox{Anim-NeRF} methods, we exclude the sections of training videos where people show their hands in front of their torso (leaving about half of the frames), this forms \textit{Scenario~A} comparison. \mbox{Anim-NeRF} limits the resolution for \textit{Scenario~A} to ${512\times512}$ px. \mbox{HF-Avatar}'s mesh fitting adequately performed for training sequences, but struggled for hold-out sequences, therefore for them we used our modified mesh fitting. \mbox{HumanNeRF} implementation does not include its own mesh fitting method, thus we use ours instead. See Table~\ref{tab:mesh-fitting} for an overview of mesh fitting methods applied in our experiments.

We compare against \mbox{HumanNeRF}, \mbox{ANR} and \mbox{StylePeople} trained on full uncut videos at $640\times640$~px resolution -- the maximum resolution at which we can run 30 FPS on a mobile device. This forms \textit{Scenario~B} comparison. For \mbox{StylePeople}, we use all the proposed components as per \cite{dnr:2019,stylepeople}, but take our segmentation, mesh fitting, losses, spectral texture initialization and architectures of the rendering and discriminator networks. In the comparison table, \mbox{StylePeople} baseline differs from \titleabbreviation{} only by the absence of the neural texture warping, in order to compare only novelty part of our work. To compare with ANR, we re-implement their split optimization scheme and add it to the \mbox{StylePeople} baseline (see supp. mat. for more details). %

To quantify visual quality, we use three metrics, computed between real and generated images. We employ two image-based metrics: SSIM~\cite{ssim:2004} and LPIPS~\cite{lpips:2018}. To measure temporal stability, we use \mbox{tLPIPS}~\cite{tlpips}.

We conduct a \textit{user study} (the right-most column of Table~\ref{tab:metrics}) in which we show pairs of sequences, one of which is the \titleabbreviation{} result (in randomized order). We took novel body motion sequences from \mbox{AMASS}~\cite{amass} dataset. We asked people to indicate their preference, and thus obtain the percentage of responses when \titleabbreviation{} was preferred over each of the other methods. 

The results are shown in Table~\ref{tab:metrics}. In our comparison, \mbox{AnimNeRF} and \mbox{InstantAvatar} produce NeRF blob artifacts. \mbox{HumanNeRF} similarly exhibits \mbox{NeRF} artifacts in motion, and its training diverged for some training videos, with the reasons unrelated to \titleabbreviation{}'s mesh fitting quality (see supp. mat. for evaluation scores and images). Due to such divergence, \mbox{HumanNeRF} has very high values for LPIPS and tLPIPS metrics. Non-NeRF \mbox{HF-Avatar} method suffers from significant blurring. In comparison to \mbox{StylePeople} and ANR, our approach outperforms both of them in temporal stability while slightly underperforming in SSIM and \mbox{LPIPS} metrics. We explain it by noticing, that when turning off warping fields on inference, the rendered avatar's texture appears in a neutral position (Fig. ~\ref{fig:warp-effect}), which is almost certainly misaligned with the real dynamic garments from the hold-out ground truth. Thus, the metrics capture mostly pixel-wise discrepancy between garments, which might not match the human perception of pleasant avatars. Qualitatively, we observe at least as good image quality for \titleabbreviation{} as that of \mbox{StylePeople} and ANR. The advantage of our method is validated by the user study on the videos, where \titleabbreviation{} is preferred over other methods. We also provide a qualitative comparison of \titleabbreviation{} with \mbox{StylePeople} in our mobile AR app in supp. mat.

In Figure~\ref{fig:methods-comparison} we show the qualitative comparisons between the methods. In Figure~\ref{fig:dnr-vs-warp}, we further compare the temporal stability of \titleabbreviation{} avatars with \mbox{StylePeople} avatars using time-space slices. Here we created avatar images with a novel animation sequence and centered them on a single \mbox{SMPL-X} mesh vertex: on a T-shirt pattern, or the hemline. We demonstrate extreme failure cases for \mbox{StylePeople} that do not occur with \titleabbreviation{}. As seen, our method provides good vertical stability of rendered parts, confirming that the tasks of mesh misalignment absorption and translation for latent neural renderings into color images are  disentangled. %

\vspace{-3mm}
\paragraph{Mesh fitting ablation study}

In Table~\ref{tab:ablation}, we perform an ablation study on mesh fitting modifications described in Section \ref{sec:details:meshfitting}. We start with the \mbox{SMPLify-X} baseline~\cite{SMPLify:2016} and add the improvements one-by-one. We retrain the \titleabbreviation{} avatars for the new fits, and perform a user study on pairs of videos: one is the baseline, the other is a modification. Every pair is shown to 25 people, one person judges at most 6 pairs every 6 hours. Although on average any modification is favored by the user study over \mbox{SMPLify-X}, on particular avatar subjects the score can significantly drop, \eg as low as 12\% for the temporal loss. However, when all modifications are used, the lowest preference score among all subjects was 52\%, indicating the consistently positive impact of the proposed fitting improvements on the avatar quality.

{\setlength{\abovecaptionskip}{3pt plus 2pt minus 2pt}
\setlength{\belowcaptionskip}{0pt plus 2pt minus 2pt}
\begin{table}[!h]
\centering
\small
\setlength{\tabcolsep}{1pt}
\caption{Ablation user study on the mesh fitting procedure. We report a percentage of users preferring avatars trained on modified mesh fits over baseline SMPLify-X~\cite{SMPLify:2016, SMPL-X:2019}.}
\rowcolors{1}{gray!10}{white}
\label{tab:ablation}
\begin{tabular}{c|c|c|c|c|c} 
\hline
 \textbf{{\small SMPLify-X}} &  \textbf{\makecell{\small+Shared\\shape}} & \textbf{\makecell{\small+Frames\\filter}} & \textbf{\makecell{\small+Temporal\\loss}} & \textbf{\makecell{\small+Silhouette\\loss}} & \textbf{\small+All}  \\
 \hline
 N/A & 57.6\% & 58.0\%  & 64.4\%  & \cellcolor{yellow!25}{84.4}\% & \cellcolor{green!25}{84.8\%}\\
 \hline
\end{tabular}
\end{table}
}

\subsection{ZJU-MoCap}

\begin{table}

\small
\setlength{\tabcolsep}{2pt} %
\newcolumntype{?}[1]{!{\vrule width #1}} %
\newcommand{\thicklinewidth}{0.4mm}
\renewcommand{\arraystretch}{1.0}

\begin{tabular}{c|c|c|c?{\thicklinewidth}c|c|c}

 & \multicolumn{3}{c?{\thicklinewidth}}{\textbf{Subject 377}}  & \multicolumn{3}{c}{\textbf{Subject 394}} \\ 
 & \multicolumn{1}{c}{\footnotesize PSNR↑} & \multicolumn{1}{c}{\footnotesize SSIM↑} & \multicolumn{1}{c?{\thicklinewidth}}{\footnotesize LPIPS↓\ensuremath{*}} & \multicolumn{1}{c}{\footnotesize PSNR↑} & \multicolumn{1}{c}{\footnotesize SSIM↑} & \multicolumn{1}{c}{\footnotesize LPIPS↓\ensuremath{*}} \\
\hline
& \multicolumn{6}{c}{ZJU-MoCap multi-view SMPL fits (type A)} \\ \cline{2-7}
HumanNeRF     & 28.54                        & 0.9557                        & \cellcolor{yellow!25}{31.203} 
              & 28.67                        & 0.9448                        & 43.980  \\ 
StylePeople   & \cellcolor{green!25}{29.30}  & \cellcolor{green!25}{0.9639}  & \cellcolor{green!25}{29.865} 
              & \cellcolor{yellow!25}{28.93} & \cellcolor{yellow!25}{0.9472} & \cellcolor{green!25}{40.828}  \\ 
\textbf{MoRF} & \cellcolor{yellow!25}{29.03} & \cellcolor{yellow!25}{0.9629} & 31.606 
              & \cellcolor{green!25}{29.13}  & \cellcolor{green!25}{0.9483}  & \cellcolor{yellow!25}{41.014}  \\ \cline{2-7}
 & \multicolumn{6}{c}{Our single-view SMPL fits (type B)} \\ \cline{2-7}
HumanNeRF     & \cellcolor{green!25}{27.74}  & \cellcolor{yellow!25}{0.9572} & \cellcolor{green!25}{38.448} 
              & \cellcolor{green!25}{27.99}  & 0.9426                        & 51.120  \\ 
StylePeople   & 27.25                        & 0.9571                        & \cellcolor{yellow!25}{42.021} 
              & \cellcolor{yellow!25}{27.92} & \cellcolor{yellow!25}{0.9440} & \cellcolor{yellow!25}{47.956}  \\ 
\textbf{MoRF} & \cellcolor{yellow!25}{27.32} & \cellcolor{green!25}{0.9573}  & 42.156 
              & 27.88                        & \cellcolor{green!25}{0.9442}  & \cellcolor{green!25}{47.692}  \\ \hline
\end{tabular}
\centering
\caption{\centering\label{tab:metrics_zju} Evaluation on ZJU-MoCap dataset. ({\footnotesize\ensuremath{*}}) LPIPS $\times 10^3$.}
\end{table}
{
\setlength{\belowcaptionskip}{-6pt}
\begin{figure}[t]
    \hspace*{-2.1em}
    \newcommand\myrowincludegraphics[2][]{\raisebox{-0.2\height}{\includegraphics[#1]{#2}}}
    \newcommand{\myfakeheight}{3.05cm}
    \newcommand{\myfakeraise}{2.1cm}
    \renewcommand\cellset{\renewcommand\arraystretch{0}}%
    \setlength\extrarowheight{0pt}
    \setlength\arrayrulewidth{-1.0em}
    \begin{tabular}{p{0.99cm}p{1.cm}p{1.01cm}p{0.95cm}p{0.5cm}p{0.9cm}} %
    
    \multicolumn{1}{c}{\footnotesize H.NeRF}
    & \multicolumn{1}{c}{\footnotesize S.People}  
    & \multicolumn{1}{c}{\footnotesize\textbf{\titleabbreviation{}}}
    & \multicolumn{1}{c}{\footnotesize H.NeRF}
    & \multicolumn{1}{c}{\footnotesize S.People}  
    & \multicolumn{1}{c}{\footnotesize\textbf{\titleabbreviation{}}}\\
    
    \myrowincludegraphics[height=\myfakeheight,trim={185 40 190 35},clip]{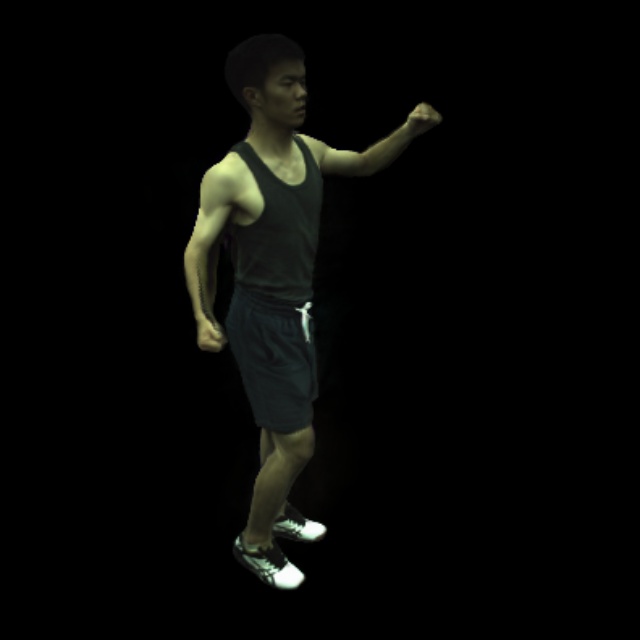}
    & \myrowincludegraphics[height=\myfakeheight,trim={190 40 185 35},clip]{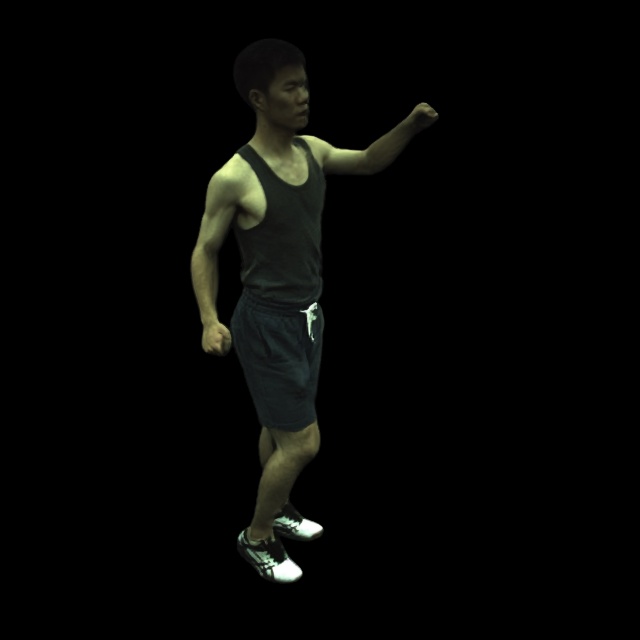}
    & \myrowincludegraphics[height=\myfakeheight,trim={187 40 181 35},clip]{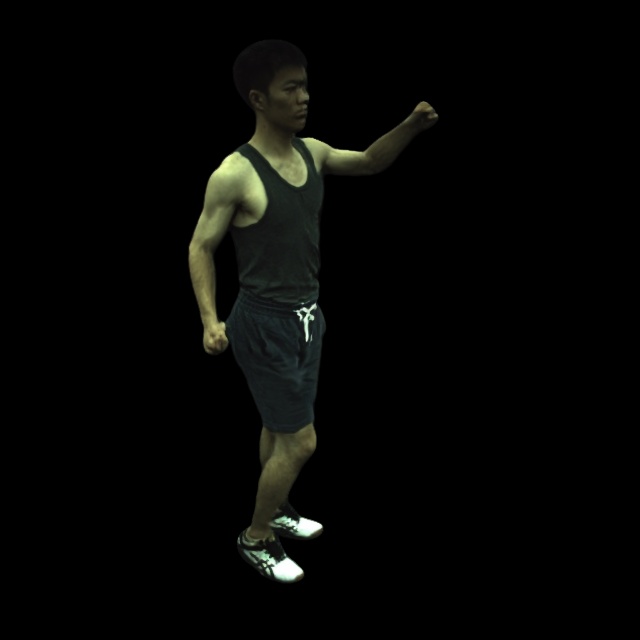}
    & \myrowincludegraphics[height=\myfakeheight,trim={193 40 180 35},clip]{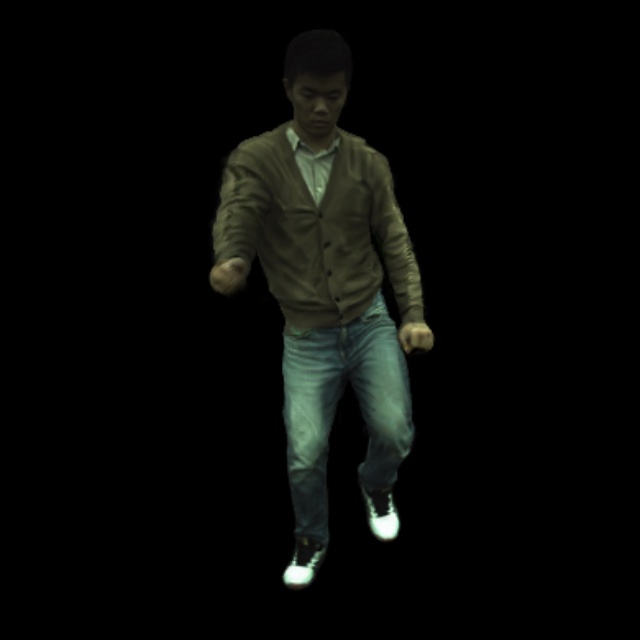}
    & \myrowincludegraphics[height=\myfakeheight,trim={195 40 183 35},clip]{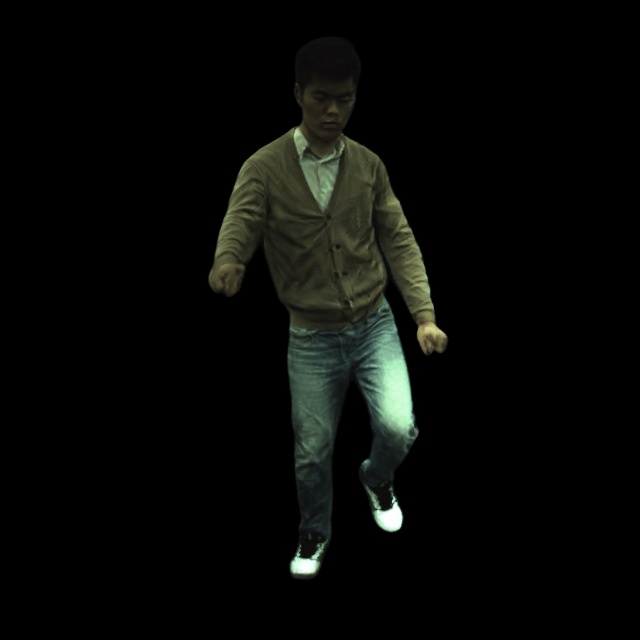}
    & \myrowincludegraphics[height=\myfakeheight,trim={195 40 190 35},clip]{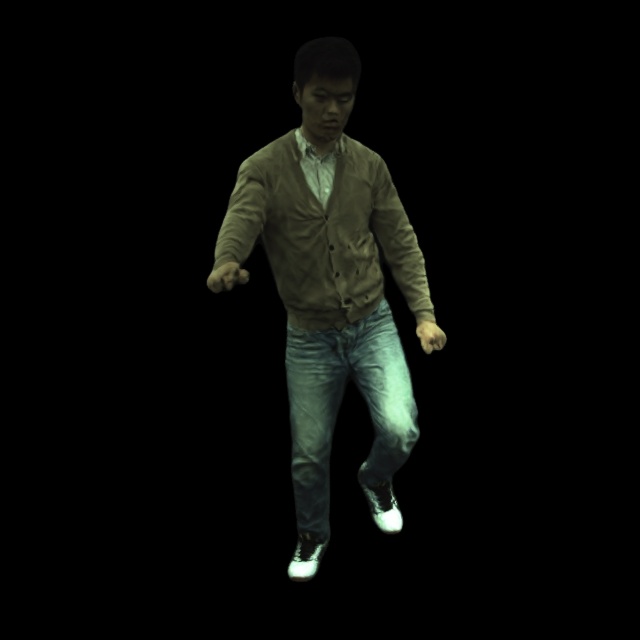}\\
    \myrowincludegraphics[height=\myfakeheight,trim={175 40 190 35},clip]{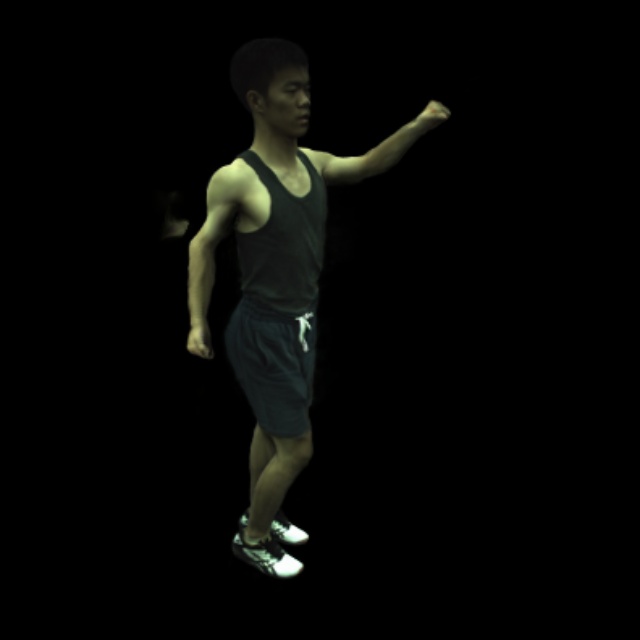}
    & \myrowincludegraphics[height=\myfakeheight,trim={190 40 185 35},clip]{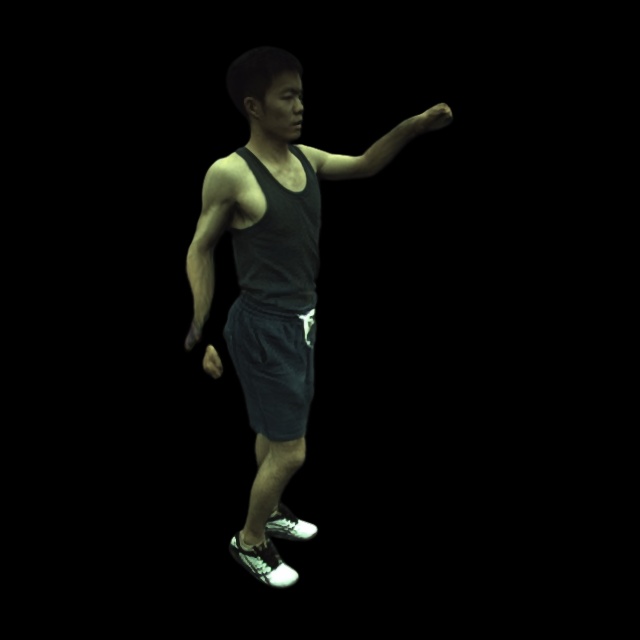}
    & \myrowincludegraphics[height=\myfakeheight,trim={190 40 150 35},clip]{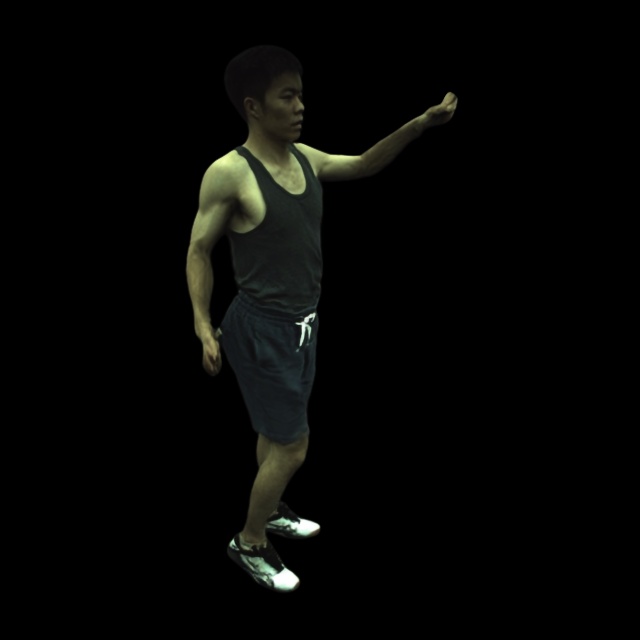}
    & \myrowincludegraphics[height=\myfakeheight,trim={193 40 180 35},clip]{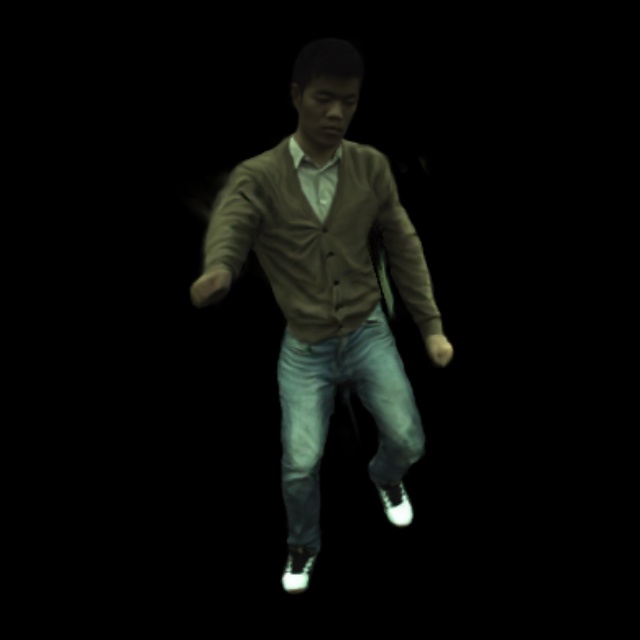}
    & \myrowincludegraphics[height=\myfakeheight,trim={193 40 185 35},clip]{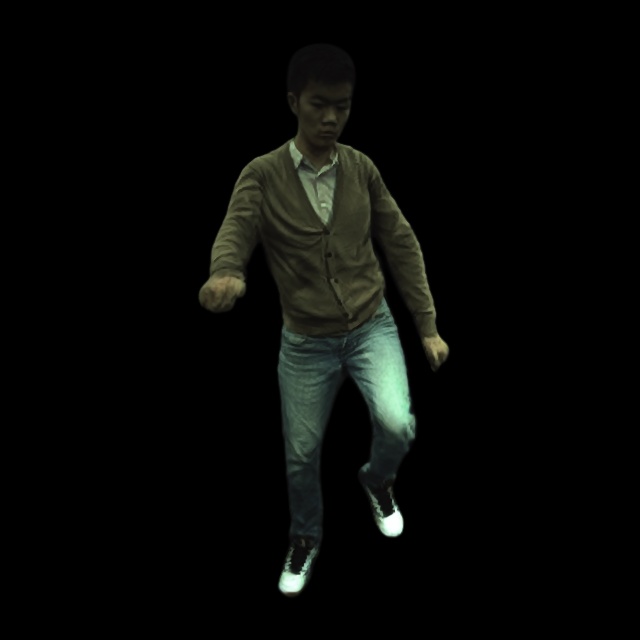}
    & \myrowincludegraphics[height=\myfakeheight,trim={190 40 195 35},clip]{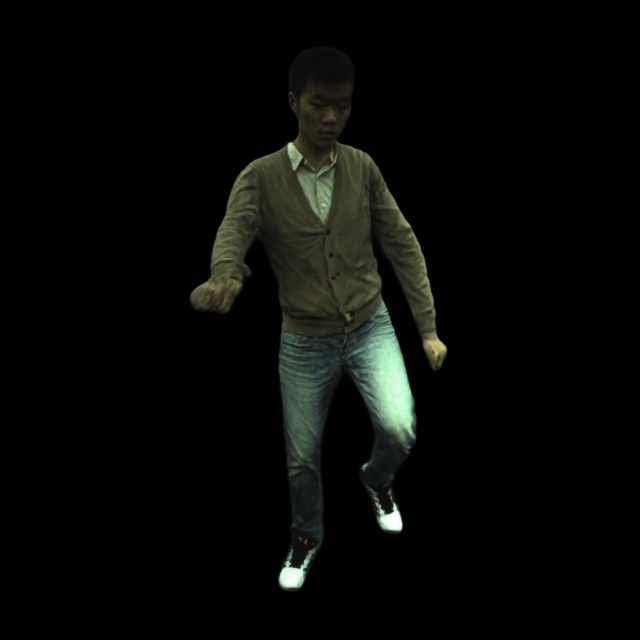}
    \end{tabular}
    \caption{ Results on ZJU-MoCap dataset (renderings of novel views far from the train view). Avatars of subjects 377 \textit{(left)}, 394 \textit{(right)} are shown, trained on multi-view \textit{(top)} or on our monocular \textit{(bottom)} mesh fits. Zoom-in is highly recommended.\label{fig:zjumocap_collage}}
\end{figure}
}
To evalute on ZJU-MoCap dataset, we follow the protocol/split from \mbox{HumanNeRF} and compare to HumanNeRF and StylePeople with two types of SMPL mesh fits: type A) the original multi-view fits from \mbox{ZJU-MoCap}, and type B) our monocular fitting results. For both types of fits, HumanNeRF exhibits ghosting and NeRF blob artifacts. Despite MoCap origin of this dataset, it's original SMPL multi-view fits (type A) are less aligned with ground truth images than our monocular fits (type B). As a result, given type A fits, we find out that MoRF performs worse that StylePeople baseline (Tab.~\ref{tab:metrics_zju}, Fig.~\ref{fig:zjumocap_collage} -- zoom-in highly recommended) because neutral texture warping could not converge. On the contrary, given well aligned type B mesh fits, MoRF produces sharper output renderings and less flicker. This sets the application scope of our rendering method: the expected effect is achieved with our mesh fitting and capture scenario.

\subsection{People Snapshot}

{
\setlength{\belowcaptionskip}{-6pt}
\begin{figure}[t]
\begin{center}
   \includegraphics[height=2.7cm]{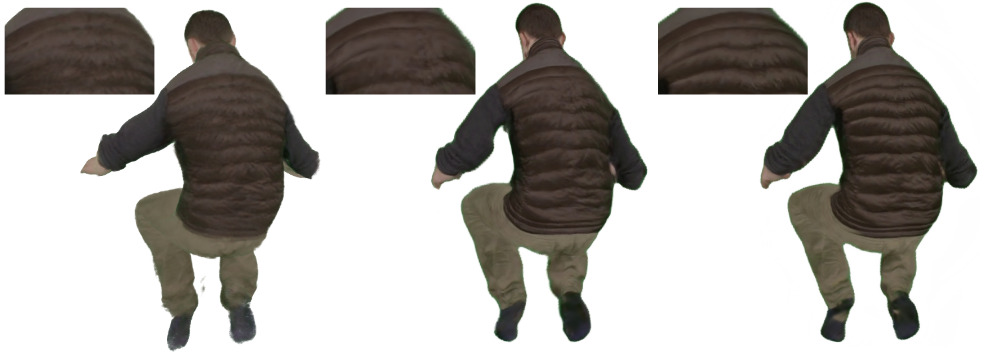}
   \includegraphics[height=2.7cm]{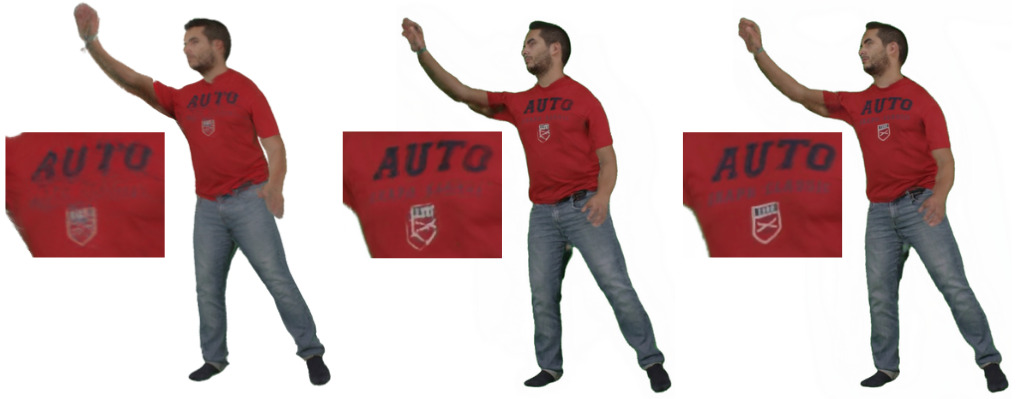}
\end{center}
   \caption{Results on PeopleSnapshot produced by InstantAvatar \textit{(left)}, StylePeople \textit{(middle)} and \titleabbreviation{} \textit{(right)}. The least texture averaging is demonstrated by \titleabbreviation{}.} 
\label{fig:people_snapshot}
\end{figure}
}

We have also found that MoRF works well on \textbf{People Snapshot} data. %
See the comparison on two idetities from People Snapshot with InstantAvatar and StylePeople in Fig.~\ref{fig:people_snapshot}.

\subsection{Experiments analyses}

For our self-captured dataset, all NeRF-based methods (\mbox{InstantAvatar}, \mbox{Anim-NeRF}  and \mbox{HumanNeRF}) produce noticeable NeRF blob artifacts, and \mbox{HumanNeRF} has the least of them. 

Compared to NeRF-based methods, mesh-based rendering methods (\mbox{StylePeople}, \mbox{ANR} and \titleabbreviation{}) are more stable and do not produce blob artifacts. Moreover the inference speed for the fastest NeRF (\mbox{InstantAvatar}) is 15 FPS for $540\times540$ px on desktop GPU, compared to 30 FPS for $640\times640$ px on \textbf{mobile} GPU for \titleabbreviation{}.

Compared to \mbox{StylePeople} and \mbox{ANR}, \titleabbreviation{} can produce sharper and more temporally stable results given our well aligned mesh fits. This effect is noticeable on texture rich regions like on a T-shift print.

For the results of \mbox{HF-Avatar} see Figure~\ref{fig:methods-comparison}. The \mbox{NeuMan} method struggles for static monocular videos like in our self-captured dataset, see supp. mat. for the exact details.

\section{Discussion}\label{sec:discussion}

We have presented a fullbody avatar system that has high realism, can be rendered in real-time on a mobile devices, and is learned from monocular videos. We evaluate our method on a self-captured dataset on novel poses, prove \titleabbreviation{} is faster and produces better rendering results compared to NeRFs. On ZJU-MoCap dataset, we find out that \titleabbreviation{} is sensitive to mesh fits quality. \titleabbreviation{} produces sharper and more temporally stable results (compared to \mbox{StylePeople} and \mbox{ANR}) given our well aligned mesh fits, the effect is noticeable on texture rich regions like on a T-shift. 

The limitations of our method are: 1) \titleabbreviation{} does not model clothing movement or loose clothing, 2) there is no prior to render train-blind areas, 3) texture seams are not handled and warping artifacts could appear there, using spherical texture without seams is a future work.

\FloatBarrier

{\small
\bibliographystyle{ieee_fullname}
\bibliography{egbib}
}

\end{document}